\documentclass[10pt,journal,compsoc]{IEEEtran}

\usepackage{times}
\usepackage{epsfig}
\usepackage{graphicx}
\usepackage{amsmath}
\usepackage{amssymb}
\usepackage{algorithm}
\usepackage{algorithmicx}
\usepackage{algpseudocode}
\usepackage{color}
\usepackage{enumitem}
\usepackage{arydshln}
\usepackage{multirow}
\usepackage{bm}
\usepackage{bbding}
\usepackage{booktabs, eucal}
\definecolor{brightcerulean}{rgb}{0., 0.46, 0.72}
\newcommand{\revise}[1]{{#1}}

\def\ie{\emph{i.e.}}
\def\eg{\emph{e.g.}}

\def\etal{{\em et al.~}}
\def\vs{{\em vs.~}}

\usepackage{hyperref}

\newlength\savedwidth
\newcommand\whline{\noalign{\global\savedwidth\arrayrulewidth
                           \global\arrayrulewidth 0.8pt}%
                  \hline
                  \noalign{\global\arrayrulewidth\savedwidth}}

\ifCLASSOPTIONcompsoc
\usepackage[nocompress]{cite}
\else
\usepackage{cite}
\fi

\ifCLASSINFOpdf

\else

\fi

\begin{document}

	\title{PAN++: Towards Efficient and Accurate End-to-End  Spotting of Arbitrarily-Shaped Text}

	\author{
		Wenhai Wang, %
		Enze Xie, %
		Xiang Li,
		Xuebo Liu,
		Ding Liang,
		Zhibo Yang,
		Tong Lu, %
		Chunhua Shen%
		\IEEEcompsocitemizethanks{
			\IEEEcompsocthanksitem W. Wang and T. Lu are with National Key Lab for Novel Software Technology, Nanjing University. Email: wangwenhai362@smail.nju.edu.cn, lutong@nju.edu.cn
			\IEEEcompsocthanksitem E. Xie is with The University of Hong Kong. Email: xieenze@hku.hk
			\IEEEcompsocthanksitem X. Li is with Nanjing University of Science and Technology. Email: xiang.li.implus@njust.edu.cn
			\IEEEcompsocthanksitem X. Liu and D. Liang are with SenseTime Group Limited. Email: \{liuxuebo, liangding\}@sensetime.com
			\IEEEcompsocthanksitem Z. Yang is with Alibaba Group. Email: zhibo.yzb@alibaba-inc.com
			\IEEEcompsocthanksitem C. Shen is with %
			Monash University, Australia.
			Email: chunhua@me.com
			\IEEEcompsocthanksitem 
			W. Wang and E. Xie contributed equally. T. Lu is the 
			corresponding author. 
		}%
	}

	\IEEEtitleabstractindextext{%
	\begin{abstract}
			Scene text detection and recognition have been well explored in the past few years. Despite the progress, efficient and accurate end-to-end spotting of arbitrarily-shaped text remains challenging.
			In this work, we propose an end-to-end text spotting framework, termed PAN++, which can efficiently detect and recognize text of arbitrary shapes in natural scenes. PAN++ is based on the kernel representation that reformulates a text line as a text kernel (central region) surrounded by peripheral pixels. By systematically comparing with existing scene text representations, we show 
			that our kernel representation can not only describe arbitrarily-shaped text but also well distinguish adjacent text. Moreover, as a pixel-based representation, the kernel representation can be predicted by a single fully convolutional network, which is very friendly to real-time applications.
			Taking the advantages of the kernel representation, we design a series of components as follows: 1) a computationally efficient feature enhancement network composed of stacked Feature Pyramid Enhancement Modules (FPEMs); 2) a lightweight detection head cooperating with Pixel Aggregation (PA); and 3) an efficient attention-based recognition head with Masked RoI. Benefiting from the kernel representation and the tailored components, our method achieves high inference speed while maintaining competitive accuracy. Extensive experiments show the superiority of our method. 
			For example, the proposed PAN++ achieves an end-to-end text spotting F-measure of 64.9 at 29.2 FPS on the Total-Text dataset, which significantly outperforms the previous best method. 
			Code will be available at: \href{https://git.io/PAN}{git.io/PAN}.

		\end{abstract}

		}

	\maketitle

	\IEEEdisplaynontitleabstractindextext
	
	\IEEEpeerreviewmaketitle

	\IEEEraisesectionheading{\section{Introduction}}
	\IEEEPARstart{R}EADING text in natural scenes is a fundamental task in numerous computer vision applications, such as text retrieval, office automation, and visual question answering.
	In virtue of the powerful representation 
	of 
	deep neural networks~\cite{simonyan2014very,he2016identity}, scene text
	detection and recognition has witnessed great progress in the past a few years~\cite{zhou2017east,psenet,crnn,shi2018aster,liu2018fots,masktextspotter}.
	Nonetheless, three main limitations still exist in these methods and hamper their deployment to real-world applications.
	
	\begin{figure}[t]
		\centering
		\begin{minipage}[b]{0.41\textwidth} 
			\centering
			\includegraphics[width=1.0\textwidth]{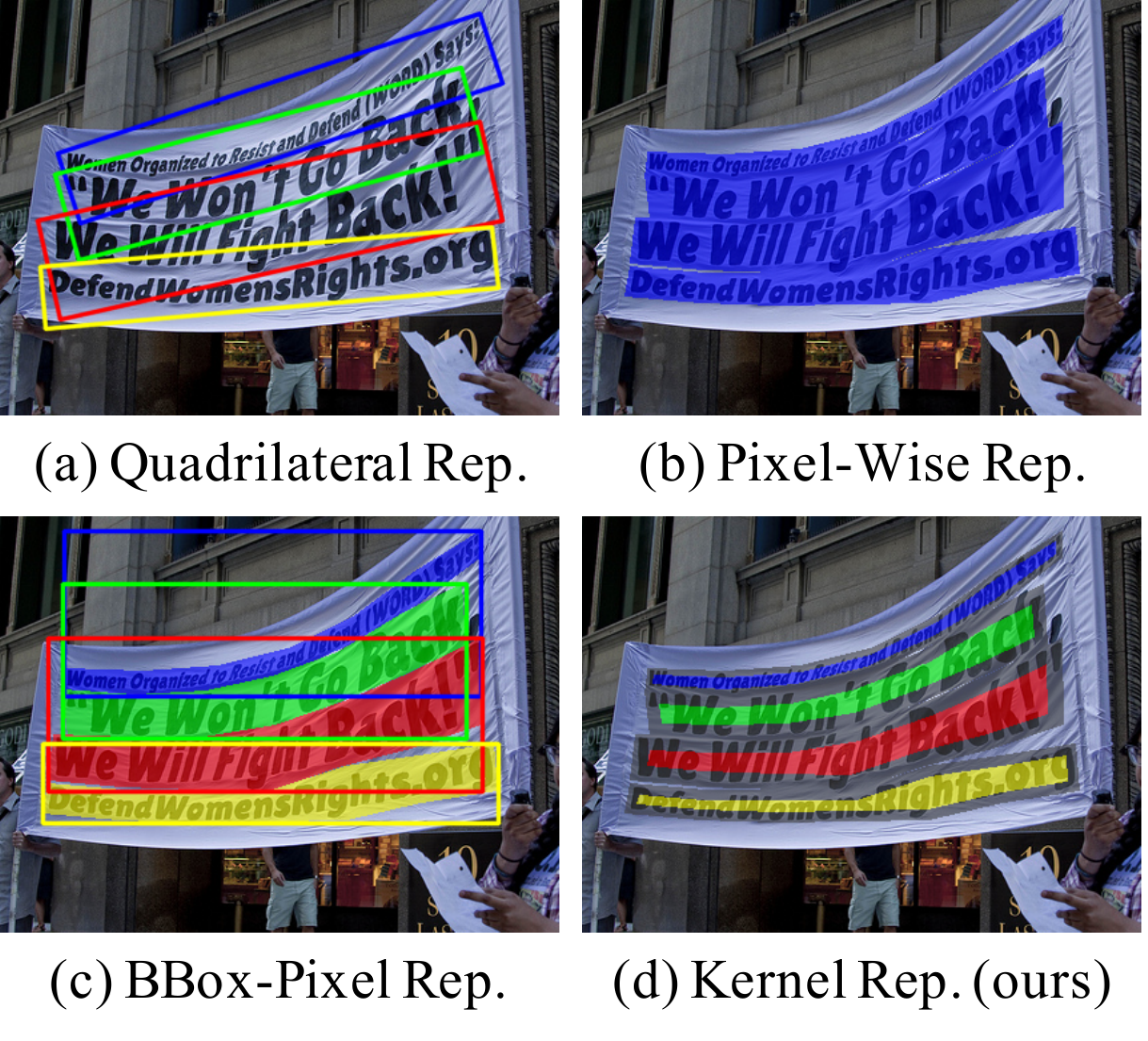}
		\end{minipage}
		\begin{minipage}[b]{0.47\textwidth} 
			\centering
			\small
			\renewcommand\arraystretch{0.85}
			\setlength{\tabcolsep}{1.7mm}
			\begin{tabular}{r|c|c|c} 
    \whline 
	Representation & Curved Text & Adjacent Text & One-Stage \\ 
	\hline
	\hline 
	Quadrilateral Rep.
& $\times$ & \checkmark & \checkmark \\ 
	\hline
	Pixel-Wise Rep.\  
& \checkmark & $\times$ & \checkmark \\ 
	\hline
	BBox-Pixel Rep.\ & \checkmark & \checkmark & $\times$ \\ 
	\hline
	Kernel Rep.\  (ours) & \checkmark & \checkmark & \checkmark \\
	\whline
\end{tabular} 
		\end{minipage}
		\caption{\textbf{Comparisons of different scene text representations.}
		(a) The quadrilateral representation~\cite{textboxes++,zhou2017east,liu2018fots}
		fail to 
		locate curved text lines. (b) The pixel-wise representation~\cite{FCN} 
		is not able to 
		separate adjacent text lines. (c) Although the bounding-box-pixel (bbox-pixel) representation~\cite{spcnet,masktextspotter,liu2018fots} can 
		accommodate  
		curved and adjacent text lines, it needs two-stage prediction, which 
		can be computationally 
		expensive. 
		(d) Our kernel representation is one-stage and  is robust to curved and adjacent text lines.}
		\label{fig:rep}
	\end{figure}
	
	First, many works tackle text detection and recognition as separate tasks by focusing on either text detection or text recognition alone.
	For most existing text detectors~\cite{tian2016detecting,zhou2017east,psenet}, a convolutional neural network is first used to generate feature maps of an input image, and then rectangular or polygonal bounding boxes of scene texts are generated using a decoder.
	On the other side, text recognition methods~\cite{crnn,shi2018aster,li2019show} often conduct a sequential prediction network on top of image patches of text lines.
	To date, few methods explore the complementarity between these two tasks. As a result, computation overhead is introduced 
	when assembling these standalone methods into a single scene text reading system.
	
	Second, most end-to-end text spotters \cite{li2017towards,liu2018fots,he2018end} are typically
	designed to read horizontal or oriented text lines.
	These methods assume that the layout of scene texts is straight, and both their detection and recognition components are developed upon this assumption.
	However, besides straight text lines, text lines with irregular character arrangements are very common in natural scenes,
	for example, texts in signboards and posters on streets.
	The methods dedicated to straight text lines %
	would fail to
	correctly detect and recognize text lines of curved shapes as shown in Fig.~\ref{fig:rep}(a).

	Last, the efficiency of existing end-to-end text spotters~\cite{li2017towards,liu2018fots,he2018end,masktextspotter,qin2019towards} remains insufficient for real-world applications.
	Although a few recent methods~\cite{masktextspotter,qin2019towards} have improved the accuracy of end-to-end arbitrarily-shaped text spotting, they suffer from low inference speed due to their heavy models or complicated pipelines.
	Thus, \emph{how to design an efficient and accurate end-to-end spotter for arbitrarily-shaped texts}
	remains largely unsolved.  
	
	\revise{
	To address the above issues, built upon 
	our preliminary results published in \cite{psenet,wang2019efficient}, here we employ the kernel representation to describe a text line by a text kernel (central region) surrounded by peripheral pixels. This is inspired by systematically studying (dis)advantages
	of the existing scene text representations  listed
	in Fig.~\ref{fig:rep}.}
	The quadrilateral representation~\cite{zhou2017east,textboxes++,liu2018fots} is the most commonly used one. %
	As shown in Fig.~\ref{fig:rep}(a), it is specifically designed 
	for straight text lines and fails to provide a tight boundary for a curved text line. 
	In contrast, 
	the pixel-wise representation proposed in \cite{FCN} is flexible enough to locate curved text lines. %
	Note that with 
	this representation, adjacent text lines %
	can 
	be conglutinated as shown in Fig.~\ref{fig:rep}(b). 
	Although the bounding-box-pixel (bbox-pixel) representation (see Fig.~\ref{fig:rep}(c)) proposed in Mask R-CNN~\cite{maskrcnn} can solve the problems mentioned above, it still follows ``detect-then-segment'' paradigm~\cite{maskrcnn}, which is often time-consuming.
	In contrast, as shown in Fig. \ref{fig:rep}(d), our kernel representation can effectively distinguish curved text lines lying closely, which is not inferior to the bbox-pixel representation. 
	Besides, 
	it can be predicted through a single convolutional network, which is conceptually simple and suitable for real-time applications.

\begin{figure*}[t]
		\centering
		\setlength{\fboxrule}{0pt}
		\fbox{\includegraphics[width=0.98\textwidth]{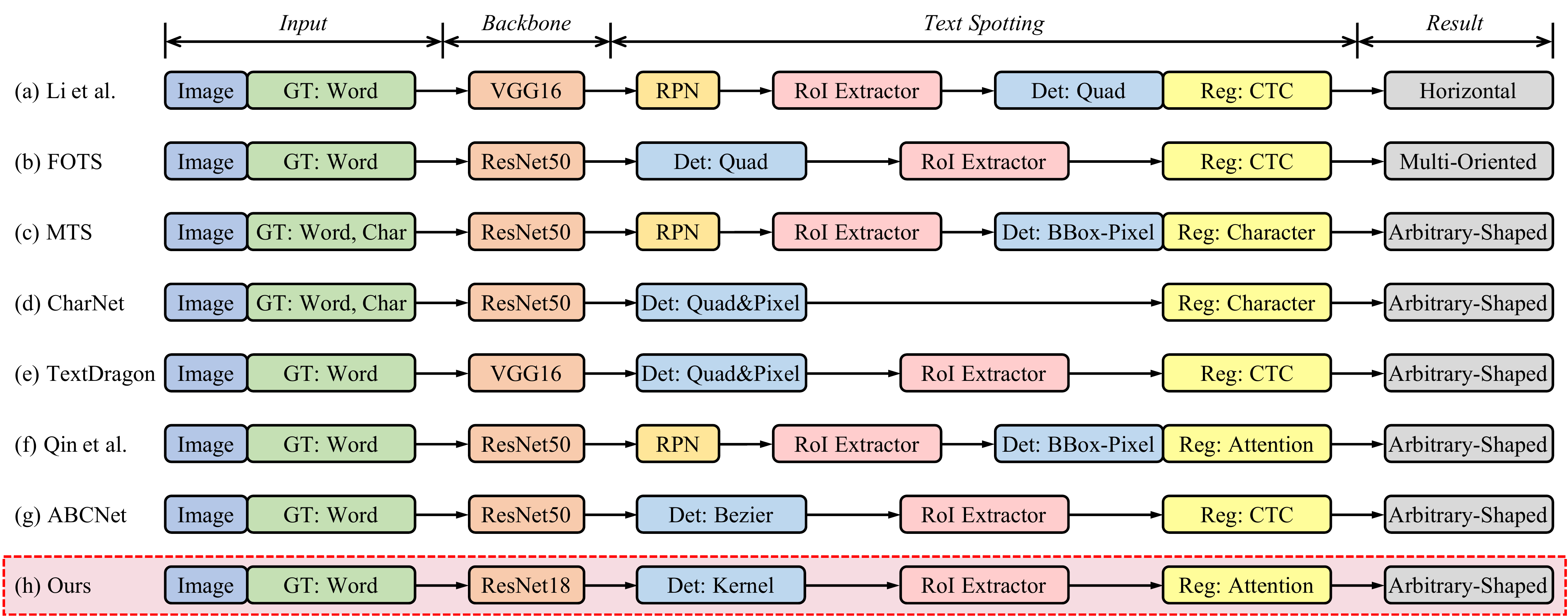}}
		\caption{\textbf{Overall pipelines of some representative text spotters relevant to ours}. In the ``GT'' (ground-truth) box, ``word'' and ``char'' represent word-level and character-level annotations, respectively.
		In the ``Det'' (Detection) box, ``Quad'', ``BBox-Pixel'', ``Kernel'' are the quadrilateral, bbox-pixel, and kernel representation mentioned in Fig.~\ref{fig:rep}.
		``Quad\&Pixel'' means performing bounding boxes regression and text region segmentation simultaneously.
		``Bezier'' represent the Bezier curve representation proposed in \cite{liu2020abcnet}.
		In ``Reg'' (recognition) box, ``CTC'', ``Character'' and ``Attention'' represent CTC-based, character-based and attention-based text recognizers, respectively. 
		}
		\label{fig:pipeline}		
	\end{figure*}

	By leveraging the advantages of the kernel representation, we further present an end-to-end arbitrarily-shaped text spotter, namely, PAN++\footnote{PAN++ is an extended version of 
	the text detector~\textbf{P}ixel \textbf{A}ggregation \textbf{N}etwork~(PAN) proposed in our conference paper~\cite{wang2019efficient}.}, 
	which can achieve a good balance between accuracy and inference speed. 
	PAN++ follows the pipeline shown in Fig.~\ref{fig:pipeline}(h), 
	which contains two main steps: 
	1) detecting text lines via a segmentation network; 
	and 2) recognizing text content via an attention-based decoder.

	To achieve high efficiency, we reduce the time cost of each step by the following four designs: 
	1) First, a lightweight backbone network~(\eg, ResNet18~\cite{he2016deep}) is employed for the segmentation network. 
	However, features extracted by %
	a 
	lightweight backbone typically have small receptive fields and weak representation capabilities.
	2) To remedy this defect, we propose a low-computation %
	feature enhancement network, which is composed of stacked Feature Pyramid Enhancement Modules (FPEMs).
	Considering
	that 
	FPEM is a U-shaped module built by separable convolutions (see Fig.~\ref{fig:fpem}), it can enhance multi-scales features with the minimal computation overhead. 
	Moreover, FPEM is stackable, which allows us to compensate for the depth of the network by stacking FPEMs to the lightweight backbone.
	3) To detect text lines, we design a simple detection head along with Pixel Aggregation (PA). The detection head predicts text regions, text kernels, and instance vectors, and PA assembles predictions of the network into complete text lines.
	4) Finally, to recognize text content, we present a feature extractor termed Masked RoI and a lightweight recognition head that contains only two LSTM~\cite{hochreiter1997long} layers and two multi-head attention layers~\cite{vaswani2017attention}.
	Benefiting from the above designs, PAN++ achieves a high inference speed while keeping competitive accuracy.
	
	\revise{
	Compared with our conference versions PSENet~\cite{psenet} and PAN~\cite{wang2019efficient}, the major extension of PAN++ lies in \emph{the text recognition module and the end-to-end text spotting framework.} In conference versions, we proposed the kernel representation and applied it to text detection.
	Different from the predecessors, we extend the text detector to an end-to-end text spotter that can fast detect and recognize arbitrarily-shaped text lines.
	To this end, we reconstruct the overall architecture of PAN++, and carefully integrate a tailored feature extractor (\ie, Masked RoI) and a lightweight text recognition head in the architecture.
	}
	
	\revise{
	Besides, 
	we also improve the text detection module when inheriting it from the conference version~\cite{wang2019efficient}.
	1) We systematically compare our kernel representation with other existing text representations.
	2) We simplify the FPEM by combining the %
	original 
	FPEM proposed in \cite{wang2019efficient} and FFM~\cite{wang2019efficient} into a single module, which is more effective.
	3) We make PA be aware of background elements by adding a background item to the discrimination loss.
	}
	
	\begin{figure}[t]
		\centering
		\setlength{\fboxrule}{0pt}
		\fbox{\includegraphics[width=0.4\textwidth]{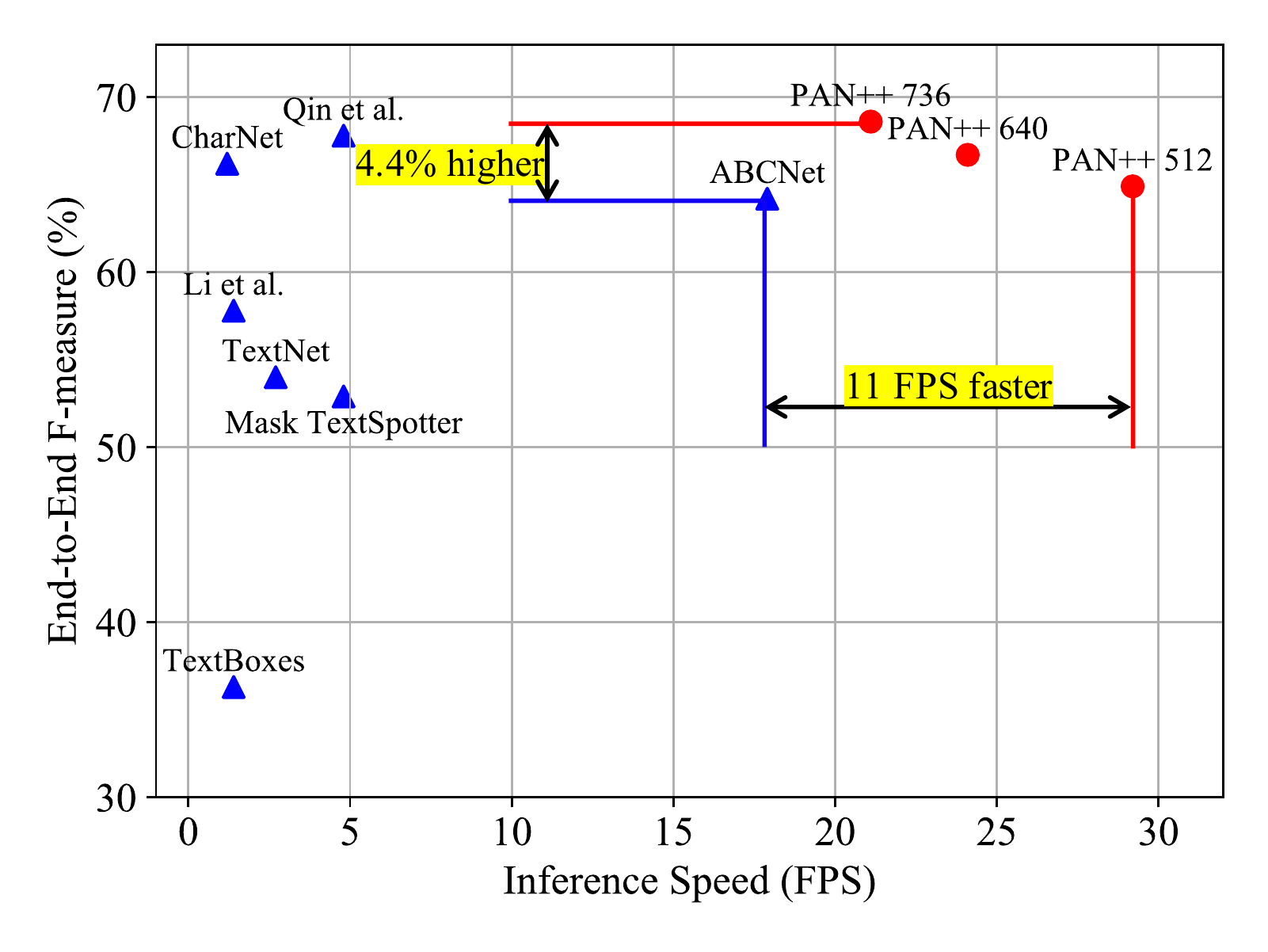}}
		\caption{\textbf{End-to-end text spotting F-measure and inference speed on Total-Text.} Our PAN++ has extreme advantages compared with counterparts. ``PAN++ 736'' (the short size of the input image being 736 pixels) is 5.0 points better than ABCNet~\cite{liu2020abcnet} and the inference speed is faster. ``PAN++ 512'' %
		executes  faster than counterparts 
		by 
		over
		11 FPS
		while keeping a competitive end-to-end text spotting F-measure.}
		\label{fig:f_fps}
	\end{figure}
	
	\revise{
	Combining these improvements, we upgrade the original text detector %
	into 
	an efficient end-to-end arbitrarily-shaped text spotter (\ie, PAN++).
	}
	To show the effectiveness of our method, we conduct extensive experiments on four challenging benchmark datasets, namely, 
	Total-Text~\cite{totaltext}, CTW1500~\cite{Liu2017Detecting}, ICDAR 2015~\cite{karatzas2015icdar}, MSRA-TD500~\cite{msra}, and RCTW-17~\cite{shi2017icdar2017}. Note that, Total-Text and CTW1500 are datasets created for curved text detection. 
	As shown in Fig.~\ref{fig:f_fps}, on the Total-Text dataset, the end-to-end text spotting F-measure of ``PAN++ 736'' reaches 68.6\%, which is 4.4 points higher than ABCNet~\cite{Liu2017Detecting}, while its inference speed is faster.
	Moreover, ``PAN++ 512'' reaches 29.2 FPS, which is 11 FPS faster than the best counterparts. At the same time, it still %
	achieves 
	a competitive end-to-end text spotting F-measure (64.9\%), which is higher than most existing methods.
	Finally, PAN++ also shows promising detection and recognition performance on other benchmarks, including multi-oriented and long text datasets.
	
	In summary, our main contributions are four-fold.
	\begin{itemize}
	     \item
	\revise{
	We %
	provide 
	the definition of the kernel representation, and
	systematically compare it with other widely-used text representations, showing that our kernel representation is conceptually simple and flexible, and friendly to real-time applications.
	}
	\item
	Based on the kernel representation, we propose a
	new framework for end-to-end arbitrarily-shaped text spotting, termed PAN++, %
	which achieves
	an excellent balance between accuracy and inference speed. 
	
	\item
	\revise{We design and improve a series of 
	efficient
	components tailored to our framework, including a feature enhancement network consisting of stacked Feature Pyramid Enhancement Modules (FPEMs), a detection head with Pixel Aggregation (PA), a feature extractor (Masked RoI), and an lightweight attention-based recognition head.}
	
	\item
	The proposed PAN++ achieves the state-of-the-art performance on curved text benchmarks, simultaneously keeping a fast inference speed. It is notable that PAN++ (\emph{S}: 512) yields an end-to-end  F-measure of 64.9\%   at 29.2 FPS on the Total-Text dataset, which is 11 FPS faster than the previous best method with competitive accuracy.

	\end{itemize}

	\section{Related Work}
	We briefly review some work most relevant to ours.
	\subsection{End-to-End Text Spotting} 
	End-to-end text spotters aim to detect and recognize text lines simultaneously in a unified network.
	In the past years, the emergence of deep learning has greatly improved the performance of text spotting.
	As shown in Fig.~\ref{fig:pipeline}, we summarize representative text spotters relevant to ours in the deep learning era. These methods can be roughly divided into two categories: 1) regular text spotter, and 2) arbitrarily-shaped text spotter.
	
	\subsubsection{Regular Text Spotter}
	Inspired by Faster R-CNN~\cite{ren2015faster}, Li \etal\cite{li2017towards} proposed the first framework for horizontal text spotting, which contains a text proposal network for text detection and a CTC-based method for text recognition, as shown in Fig.~\ref{fig:pipeline}(a).
	Meanwhile, Busta \etal\cite{busta2017deep} designed a similar framework to the work in \cite{li2017towards}. But its detection head works better for both horizontal and multi-oriented text instances.
	\revise{
	NguyenVan \etal\cite{nguyenvan2019pooling} developed an end-to-end text reading framework by incorporating a pooling-based scene text proposal technique, where false alarms elimination and words recognition are performed simultaneously.
	}
	Subsequently, Liu \etal\cite{liu2018fots} proposed a multi-oriented text spotter shown in Fig.~\ref{fig:pipeline}~(b), termed FOTS, which is equipped with a new RoI extractor~(\ie, RoIRotate) to extract the features of quadrilateral text instances.
	A similar framework is also developed by He \etal\cite{he2018end}, whose recognition head is implemented by an attention-based sequence-to-sequence decoder.
	Although these methods have achieved good performance across straight text benchmarks (\eg, IC15~\cite{karatzas2015icdar} and MSRA-TD500~\cite{msra}), they fail to detect and recognize text lines of curved shapes (see Fig.~\ref{fig:rep}(a)).

	\subsubsection{Arbitrarily-Shaped Text Spotter}
	\label{sec:re_ats}
	
	As illustrated in Fig.~\ref{fig:pipeline}(c), Liao \etal\cite{masktextspotter} proposed Mask TextSpotter (MTS) by adding character-level supervision to Mask R-CNN~\cite{maskrcnn} to detect and recognize text lines and characters simultaneously.
	This is probably the first arbitrarily-shaped text spotter.
	Note that character-level annotations are not always available. 
	The improved version~\cite{liao2019mask} of Mask TextSpotter removed the dependence of character-level annotations.
	CharNet in \cite{xing2019charnet} is the first one-stage arbitrarily-shaped text spotter as shown in Fig.~\ref{fig:pipeline}(d). Similar to Mask TextSpotter, it requires character-level annotations for training.
	Feng \etal\cite{feng2019textdragon} proposed TextDragon as shown in Fig.~\ref{fig:pipeline}(e), which contains a novel RoI extractor (\ie, RoISlide) to represent a text line by features of multiple text segments.
	Qin \etal\cite{qin2019towards} designed an arbitrarily-shaped text spotter as shown in Fig.~\ref{fig:pipeline}(f), where an attention-based text recognition head was added on Mask R-CNN to simultaneously detect and recognize irregular text lines.
	Recently, Liu \etal\cite{liu2020abcnet} formulated irregular text instances with  parameterized Bezier curves, and proposed a regression-based arbitrarily-shaped text spotter ABCNet~(see Fig.~\ref{fig:pipeline}(g)).
	Note that pixel-based (mask) representation used in the previous methods~\cite{psenet,masktextspotter,qin2019towards} may be
	more flexible and reliable than the Bezier curve representation due to the fact that the latter one can only represent a series of curves under specific constraints.

	\subsubsection{Real-time Arbitrarily-Shaped Text Spotter}
	In \cite{liu2018fots}, Liu \etal proposed a fast text spotter FOTS to detect and recognize multi-oriented texts. However, FOTS cannot 
	process 
	curved text lines.
	Most methods~\cite{masktextspotter,liao2019mask,xing2019charnet,feng2019textdragon,qin2019towards} mentioned in Sec.~\ref{sec:re_ats} have achieved high accuracies on arbitrarily-shaped text spotting, but the inference speed is rarely addressed in those methods. 
	
	This inspires us to design an improved pixel-based representation and develop an effective end-to-end text spotting framework based on it in this work.
	
	\begin{figure*}[t]
		\centering
		\setlength{\fboxrule}{0pt}
		\fbox{\includegraphics[width=0.95\textwidth]{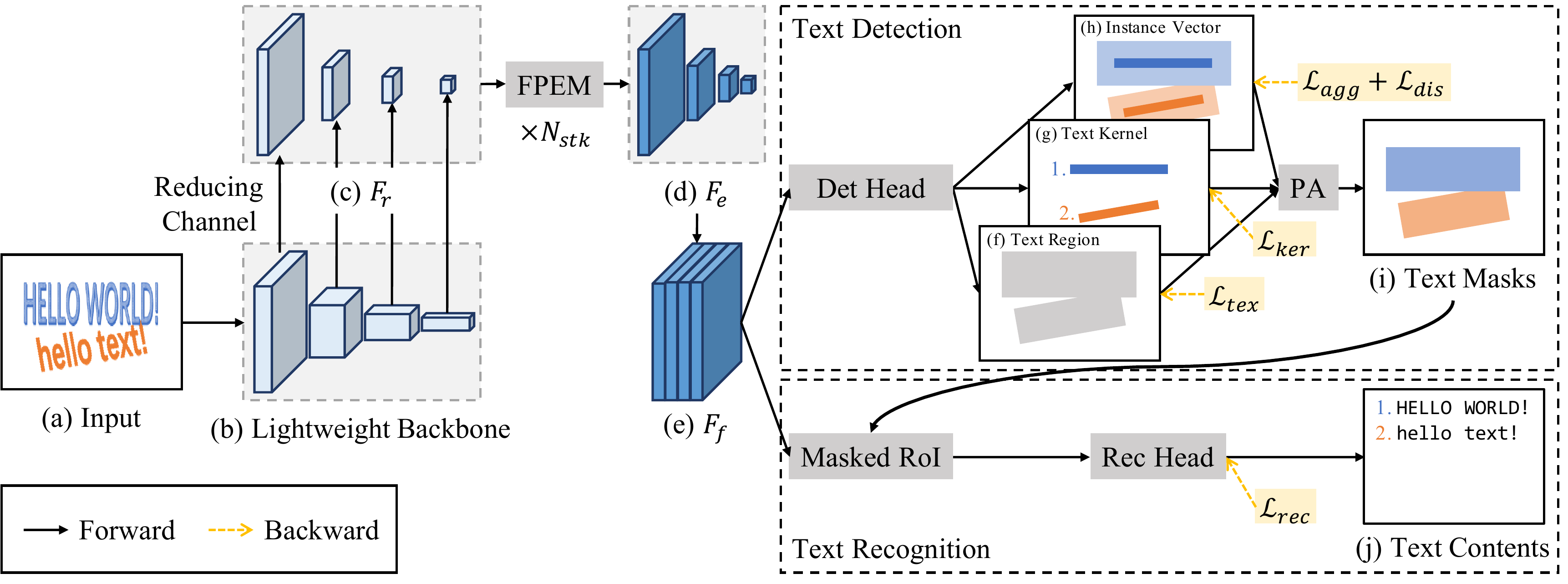}}
		\caption{\textbf{Overall architecture of PAN++}. The features from the lightweight backbone network are enhanced by stacked Feature Pyramid Enhancement Modules (FPEMs). In the detection part, the network predicts the text regions, text kernels, and instance vectors, and assembles them into text lines via Pixel Aggregation (PA). In the recognition part, Masked RoI extracts the features of text lines, and then these features are fed to the attention-based recognizer to predict text contents.
		}
		\label{fig:arch}
	\end{figure*}
	
	\subsection{Scene Text Detection}
	Scene text detection is a critical part of text spotting systems.
	Recently, methods based on deep learning have become the mainstream of text detection.
	Tian \etal\cite{tian2016detecting} and Liao \etal\cite{liao2017textboxes} successfully adapted object detection frameworks for text detection, and achieved good performance on horizontal text datasets. 
	Then researchers took orientations of text instances into consideration and developed different methods~\cite{zhou2017east,shi2017detecting,textboxes++} to detect multi-oriented text instances. 
	Most of these methods make text detection by regressing quadrilateral bounding boxes, failing to detect curved text lines.
	\revise{
	To detect curved text lines, some early works~\cite{tian2015text,shi2017detecting} followed a bottom-up framework to first detect characters or text fragments and then link them. Recently, some researchers designed segmentation-based methods~\cite{PixelLink,psenet,wang2019efficient} or combined the advantages of segmentation and box/point regression~\cite{textsnake,spcnet,xue2019msr}, achieving excellent performances on curved text detection benchmarks. 
	}
	These methods were designed to solve text detection alone and typically text recognition is achieved using a separate model.
	
	\revise{
	In this work, we not only improve the performance of text detection by summarizing the advantages of our conference versions~(\ie, PSENet~\cite{psenet} and PAN~\cite{wang2019efficient}), 
	but also carefully integrate a recognition module in our framework to build a unified model for text spotting.
	}
	
	\subsection{Scene Text Recognition}
	Scene text recognition is the last step in text spotting systems, which aims to decode sequence of characters from the visual input.
	Existing text recognizers can be roughly split into three categories, namely, 1) character-based text recognizers, 2) 
	CTC~\cite{graves2006connectionist}-based text recognizers, and 3) attention-based text recognizers.
	Character-based text recognizers \cite{bissacco2013photoocr,jaderberg2014deep} mostly first detect individual characters and then group them into words. 
	\revise{CTC-based methods~\cite{su2014accurate,su2017accurate,crnn,he2016reading} often stack RNN on top of CNN to capture long-range sequence features.}
	These methods are trained with CTC loss~\cite{graves2006connectionist}, and their the final prediction are produced by removing duplicated outputs in the testing phase.
	In attention-based methods~\cite{yang2017learning,cheng2018aon,shi2018aster,li2019show}, RNN or Transformer~\cite{vaswani2017attention} is used to predict a character at each step based on the features from attention mechanisms and previous steps.
	The process will stop when it predicts the end-of-sequence (EOS) or reaches the maximum number of steps.
	To recognition irregular text lines, \revise{Shi \etal\cite{shi2018aster} and Zhan \etal\cite{zhan2019esir} 
	used 
	a spatial rectification network to rectify irregularly-shaped text lines into straight ones before recognition.} Besides that, some recent works~\cite{yang2017learning,li2019show} applied the attention mechanism on two-dimensional feature maps to recognize texts with irregular shapes.
	
	Inspired by \cite{li2019show} and \cite{vaswani2017attention}, we design a lightweight attention-based recognition head for irregular text recognition by utilizing multi-head attention to fuse visual and temporal features from CNN and RNN, respectively.

	\section{Proposed Method}
	\subsection{Kernel Representation}
	\label{sec:kernel_rep}
	As presented in Fig.~\ref{fig:rep} (d), the kernel representation formulates a text line as a text kernel surrounded by peripheral pixels. In other words, for each given text line, we first locate it through the text kernel (the central region of the text line). Then, we recover the complete shape of the text line by involving text pixels around the text kernel.
	
	In general, there are four advantages of our kernel representation. 
	1) As a pixel-based representation, it is flexible enough to represent text lines of arbitrary shapes; 
	2) Because there are large geometrical margins among text kernels, the kernel representation can accurately distinguish adjacent text instances, solving the conglutination problem caused by adjacent text lines shown in Fig.~\ref{fig:rep} (b);
	3) The kernel representation is totally pixel-based, meaning that
	it can be easily predicted by a single fully convolutional network, which is friendly to real-time applications.
	4) The label of the kernel representation can be simply generated without additional annotations, as shown in Fig.~\ref{fig:label_gen}.
	
	\subsection{Overall Architecture}
	Based on the kernel representation, we develop an efficient text spotting framework termed PAN++, whose architecture is illustrated in Fig.~\ref{fig:arch}.
	For high efficiency, the backbone network (CNN) should be lightweight, such as ResNet18.
	However, the features produced by the lightweight backbone network often have small receptive fields and weak representation capabilities.
	To handle this problem, we propose a feature enhancement network that can refine the features efficiently. It consists of several stacked Feature Pyramid Enhancement Modules~(FPEMs).
	As shown in Fig.~\ref{fig:fpem}, FPEM is stackable and computationally efficient, which 
	can be attached behind the backbone network and makes features deeper and more expressive than before.
	For text detection, we propose a simple detection head with Pixel Aggregation (PA), which predicts: 1) text regions to describe the complete shapes of text lines, 2) text kernels to distinguish different text lines, and 3) instance vectors to recover complete text lines from text kernels, and then these predictions are combined into the text line by PA.
	For text recognition, Masked RoI is employed to extracted feature patches of text lines, and an attention-based recognition head is proposed to recognize text content.
	
	In the inference phase, we firstly feed an input image of a size of $H \times W \times 3$ to a lightweight backbone network (\eg, ResNet18~\cite{he2016deep}).
	Four feature maps~(see Fig.~\ref{fig:arch}~(b)) are generated by conv2, conv3, conv4, and conv5 stages of the backbone network, 
	whose resolutions are 1/4, 1/8, 1/16 and 1/32 compared with the input image, respectively.
	Then, we reduce the channel number of each feature map to 128 via 1$\times$1 convolutions, and obtain a thin feature pyramid $F_r$ (see Fig.~\ref{fig:arch}~(c)).
	The feature pyramid is enhanced by $N_{stk}$ stacked FPEMs.
	After 
	obtaining 
	enhanced feature pyramid $F_e$ (see Fig.~\ref{fig:arch}~(d)), we upsample and concatenate the feature maps of feature pyramid $F_e$ into the final feature map for follow-up text detection and recognition.
	The final feature map is termed $F_f$ (see Fig.~\ref{fig:arch}~(e)), whose size is $H/4 \times W/4 \times 512$.
	Next, based on the feature map $F_f$, the detection head predicts text regions, text kernels, and instance vectors, and then Pixel Aggregation (PA) assembles them into complete text lines (see Fig.~\ref{fig:arch}~(i)).
	During text recognition, we first reduce the channel number of feature map $F_f$ to 128 by a 3$\times$3 convolution and apply Masked RoI to extract the feature patches according to the predicted text lines.
	Finally, using the extracted feature patches as input, the recognition head recognizes the text content in each patch, as shown in Fig.~\ref{fig:arch} (j).
	
	During training, we use loss functions $\mathcal{L}_{tex}$,  $\mathcal{L}_{ker}$, $\mathcal{L}_{agg}$, and $\mathcal{L}_{dis}$ to optimize the predicted text regions, text kernels and instance vectors, respectively. 
	At the same time, the loss function $\mathcal{L}_{rec}$ is applied to optimize the prediction of the recognition head. 
	To maintain the consistency of recognition features, we use ground-truth bounding boxes to extract feature patches in the training phase.
	
	\subsection{Feature Pyramid Enhancement Module}
	\begin{figure}[t]
		\centering
		\setlength{\fboxrule}{0pt}
		\fbox{\includegraphics[width=0.48\textwidth]{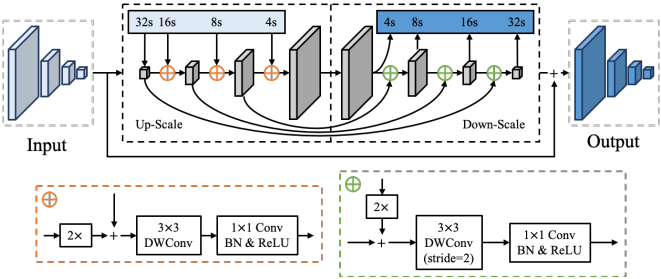}}
		\caption{\textbf{Details of the Feature Pyramid Enhancement Module (FPEM)}. ``$+$'' means element-wise addition. ``2$\times$'' indicates 2$\times$ bilinear upsampling.``DWConv'', ``Conv'' and ``BN'' represent, depthwise convolution~\cite{howard2017mobilenets}, regular convolution~\cite{lecun1998gradient} and Batch Normalization~\cite{ioffe2015batch}, respectively.}
		\label{fig:fpem}
	\end{figure}
	The Feature Pyramid Enhancement Module (FPEM) is the basic unit of the feature enhancement network, which is U-shaped as presented in Fig.~\ref{fig:fpem}. 
	FPEM consists of two phases, including up-scale enhancement and down-scale enhancement phases. 
	The up-scale enhancement is applied to the input feature pyramid, which enhances the input feature maps with strides of 32, 16, 8, and 4 pixels iteratively.
	In the down-scale enhancement phase, the input is the feature pyramid generated by up-scale enhancement, and the enhancement is conducted from 4-stride to 32-stride iteratively.
	Finally, the output feature pyramid is the element-wise addition result of the input feature pyramid and the feature pyramid produced by the down-scale enhancement.
	
	As shown in the dashed boxes in Fig.~\ref{fig:fpem}, we employ the separable convolution~\cite{howard2017mobilenets} (a 3$\times$3 depthwise convolution followed by a 1$\times$1 projection) instead of the regular convolution to implement the join part $\oplus$ of FPEM.
	Therefore, FPEM is capable of enlarging the receptive field (thanks to 3$\times$3 depthwise convolutions) and deepening the network (thanks to 1$\times$1 projections) with a small computation overhead.
	
	Similar to FPN~\cite{lin2017feature}, FPEM can enhance multi-scale feature maps by fusing the low-level and high-level information.
	Beyond that, there are two other advantages of FPEM.
	1) FPEM is a stackable module. With the increment of stack number $N_{stk}$, the feature maps will be more adequately integrated, and their receptive fields will become larger.
	2) FPEM is computationally efficient. Benefiting from separable convolutions, FPEM only involve marginal computational overhead.
	As reported in Table~\ref{tab:abs_pspnet}, the model equipped with FPEMs runs 8.7 FPS faster than the model with FPN on the IC15 dataset, while keeping a higher end-to-end text spotting F-measure.
	
	\subsection{Text Detection}
	\subsubsection{Detection Head}
	\begin{figure}[t]
		\centering
		\setlength{\fboxrule}{0pt}
		\fbox{\includegraphics[width=0.48\textwidth]{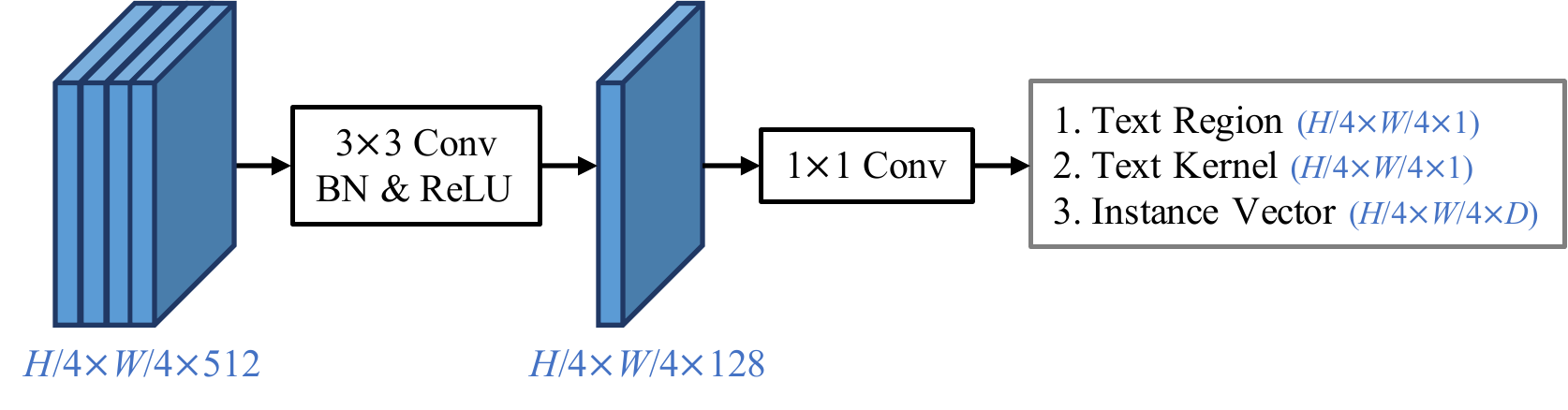}}
		\caption{\textbf{Details of the detection head}. ``Conv'' and ``BN'' represent regular convolution~\cite{lecun1998gradient} and Batch Normalization~\cite{ioffe2015batch}, respectively.}
		\label{fig:det}
	\end{figure}
	The detection head of PAN++ consists of only two convolutions, as shown in Fig.~\ref{fig:det}. The outputs of the detection head are text regions, text kernels, and instance vectors.
	1) Text regions keep the complete shape of text lines, but closely lying text lines are often overlapping (see Fig.~\ref{fig:arch} (f)).
	2) 
	In contrast, 
	text kernels can easily distinguish adjacent text instances, but they lose the complete shape of text lines (see Fig.~\ref{fig:arch} (g)).
	\revise{3) The instance vector is a high-dimensional representation that contains instance information of each pixel. In other words, pixels belonging to the same text line tend to have similar instance vectors (see Fig.~\ref{fig:arch} (h)).
	Therefore, instance vectors can direct the pixels in text regions towards the corresponding text kernel, which successfully combines the advantages of text regions and text kernels.
	}

	The three outputs complement each other and make it possible to describe text lines of arbitrary shapes.
	
	\subsubsection{Pixel Aggregation}
	\begin{figure}[t]
		\centering
		\setlength{\fboxrule}{0pt}
		\fbox{\includegraphics[width=0.45\textwidth]{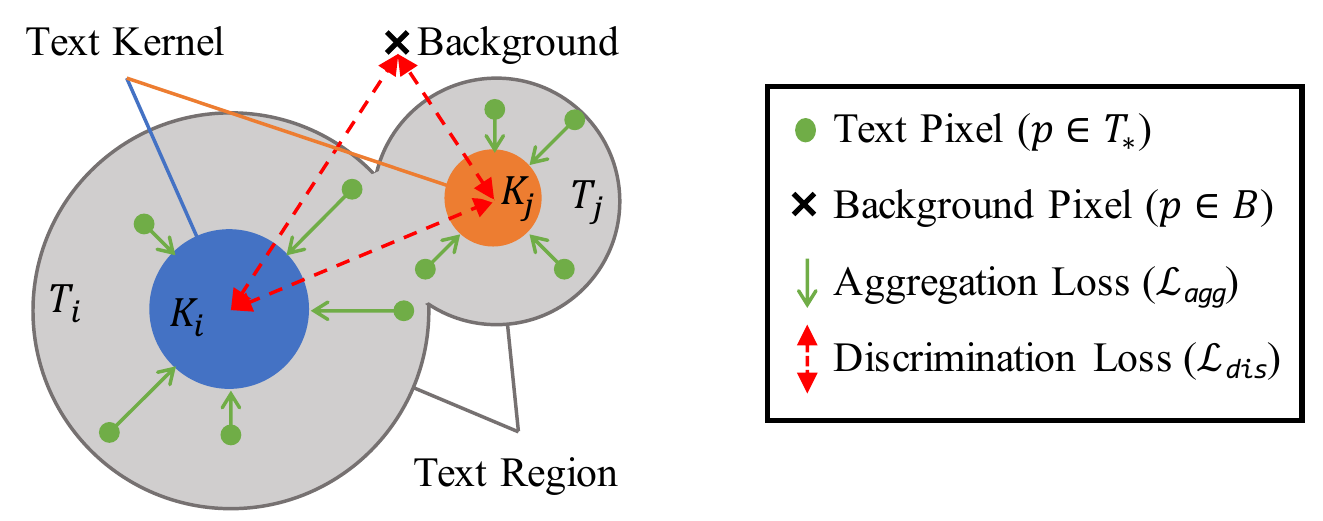}}
		\caption{\revise{\textbf{Illustration of Pixel Aggregation (PA)}. Green arrows represent aggregation loss $\mathcal{L}_{agg}$. Red arrows represent discrimination loss $\mathcal{L}_{dis}$.}}
		\label{fig:dis}
	\end{figure}
	Pixel aggregation (PA) is applied to optimize instance vectors and assemble the outputs of the detection head.
	We borrow the idea of clustering to design PA.
	As shown in Fig.~\ref{fig:dis}, if we treat different text lines as different clusters, their text kernels are cluster centers, and the pixels inside text regions are the samples the samples to be clustered.
	
	Naturally, to cluster pixels to the corresponding text kernels (see the green arrows in Fig.~\ref{fig:dis}), the distance between the pixel and the text kernel inside the same text lines should be minimized. 
	In the training phase, we use an aggregation loss $\mathcal{L}_{agg}$ formulated as Equ.~\ref{eqn:loss_agg} to implement this rule.
	\begin{equation}
		\mathcal{L}_{agg} = \frac{1}{N} \sum_{i=1}^{N} \frac{1}{\left| T_i \right|} \sum_{p \in T_i} \mathcal{D}_1(p, K_i),
		\label{eqn:loss_agg}
	\end{equation}
	\begin{equation}
		\mathcal{D}_1(p, K_i) = {\rm ln} (\mathcal{R}(\| \mathcal{F}(p) - \mathcal{G}(K_i) \| - \delta_{agg})^2 + 1).
		\label{eqn:dis_pk}
	\end{equation}
	\revise{
	Generally, the aggregation loss $\mathcal{L}_{agg}$ is employed to pull the text pixel to the target text kernel. The symbols in Eqns.~\eqref{eqn:loss_agg} and \eqref{eqn:dis_pk} are defined as follows:
	\begin{itemize}
	\item $N$: the number of text lines. 
	\item $T_i$: the text region of the $i$th text line. 
	\item $K_i$: the text kernel $K_i$ of the text line $T_i$.
	\item $\mathcal{D}_1(p, K_i)$: the distance between the text pixel $p$ and the text kernel $K_i$.
	\item $\mathcal{R}(\cdot)$: the ReLU function, ensuring the output is non-negative.
	\item $\mathcal{F}(p)$: the instance vector of the pixel $p$.
	\item $\mathcal{G}(K_i)$: the instance vector of the text kernel $K_i$, which can be calculated by: $\mathcal{G}(K_i) = \sum_{p \in K_i}\mathcal{F}(p) / \left| K_i \right|$.
	\item $\delta_{agg}$: a constant, which is set to 0.5 experimentally.
	\end{itemize}
	}
	
	\revise{
	Besides that, the cluster centers need to keep discrimination correspondingly. As a result, the instance vector of the text kernel should keep a certain distance from other text kernels and the background, as shown in the red arrows in Fig.~\ref{fig:dis}.
	This rule can be summarized as a discrimination loss $\mathcal{L}_{dis}$ as follows:
	}
	\begin{equation}
		\mathcal{L}_{dis} = \frac{1}{N^2} \sum_{i=1}^{N} (D_{b}(K_i) + \mathop{\sum_{j=1}^{N}}\limits_{j\neq i} \mathcal{D}_2(K_i, K_j)),
		\label{eqn:loss_dis}
	\end{equation}
	\begin{equation}
		\mathcal{D}_{b}(K_i) = \frac{1}{\left|B\right|}\sum_{p\in B}{\rm ln}(\mathcal{R}(\delta_{dis} - \left\| \mathcal{F}(p) - \mathcal{G}(K_i) \right\|)^2 + 1),
		\label{eqn:dis_b}
	\end{equation}
	\begin{equation}
		\mathcal{D}_2(K_i, K_j) = {\rm ln}(\mathcal{R}(\delta_{dis} - \left\| \mathcal{G}(K_i) - \mathcal{G}(K_j) \right\|)^2 + 1).
		\label{eqn:dis_kk}
	\end{equation}
	\revise{
	The goal of the discrimination loss $\mathcal{L}_{dis}$ is to push the text kernels and the background away from each other.
	The following is the definition of the symbols in  Eqns.~\eqref{eqn:loss_dis}, \eqref{eqn:dis_b}, and \eqref{eqn:dis_kk}.
	\begin{itemize}
	    \item $B$: the background.
	    \item $\mathcal{D}_b(K_i)$: the distance between the text kernel $K_i$ and the background.
	    \item $\mathcal{D}_2(K_i, K_j)$: the distance between the text kernel $K_i$ and the text kernel $K_j$.
	    \item $\delta_{dis}$: a constant, which is set to 3 in all experiments.
	\end{itemize}
	}
	
\begin{figure}[t]
		\centering
		\setlength{\fboxrule}{0pt}
		\fbox{\includegraphics[width=0.48\textwidth]{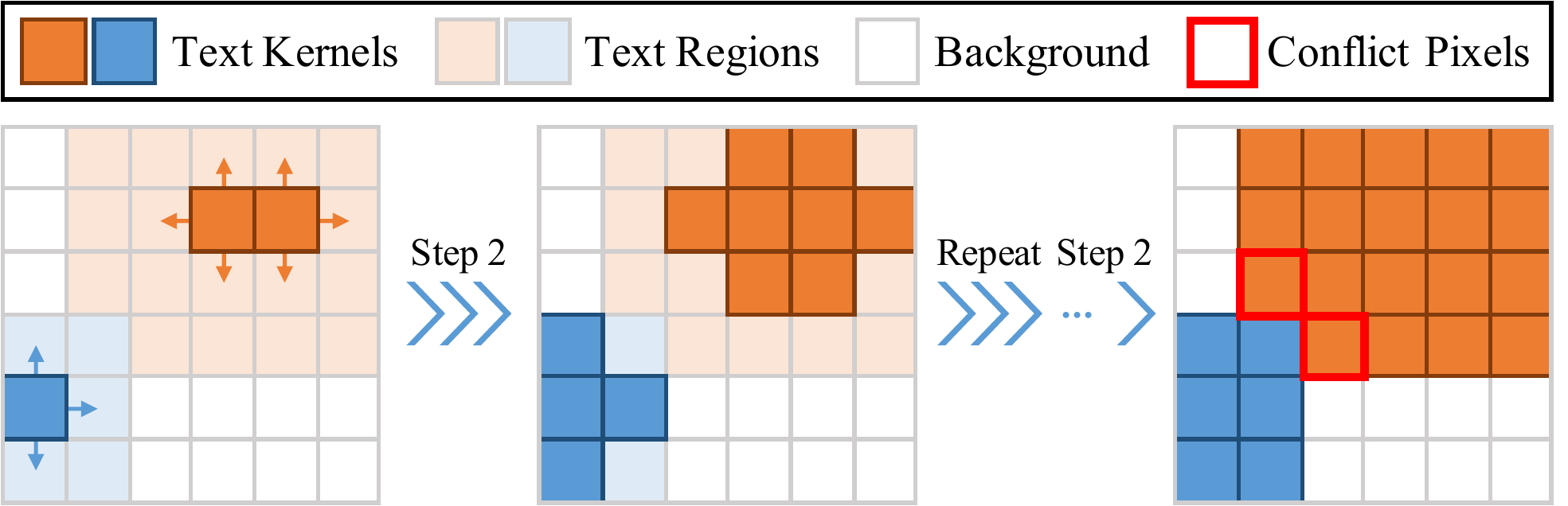}}
		\caption{\textbf{Procedure of the post-processing}. When merging neighbor pixels, the Euclidean distance of the instance vector must be less than a threshold $d$. This condition alleviates the problem of assigning conflict pixels to incorrect text kernels.}
		\label{fig:dis_infer}
		
	\end{figure}

	In the testing phase, we utilize the predicted instance vector to guide the pixels in text regions to the corresponding text kernel. As shown in Fig.~\ref{fig:dis_infer}, the post-processing has three steps as follows: 
	1) finding the connected components in the segmentation result of text kernels, and each connected component is regarded as a text kernel; 
	2) for each text kernel $K_i$, involving its neighbor pixel $p$ (4-way) in text regions when the Euclidean distance of their instance vectors is less than $d$;
	3) repeating the second step until there is no available neighbor pixel in text regions.
	
	\subsection{Text Recognition}
	\subsubsection{Masked RoI}
	Masked RoI is an RoI extractor used to extract the fixed-size feature patches for arbitrarily-shaped text lines. It contains four steps:
	1) calculating the minimal upright bounding rectangle containing the target text line;
	2) extracting the feature patch within the upright bounding rectangle.
	3) filtering noise features by multiplying the feature patch by a binary mask, where the weight outside the target text line is 0.
	4) resizing the feature patch to a fixed size. These steps can be summarized as Eqn.~\eqref{eqn:mroi}.
	\begin{equation}
		F_{roi} = {\rm Resize}_{h\times w}({\rm Crop}(F_f, {\rm Rect}(m_i)) \circ \ m_i),
		\label{eqn:mroi}
	\end{equation}
	where $m_i$ is the binary mask of the $i$th text line.
	${\rm Rect(\cdot)}$ refers to the upright bounding rectangle of a binary mask. 
	${\rm Crop}(F_f, \cdot)$ is the operation of cropping the feature patch inside an upright RoI region from the feature map $F_f$. $\circ$ means element-wise multiplication. 
	${\rm Resize}_{h\times w}(\cdot)$ represents the operation of resizing the feature map's size to $h\times w$ via bilinear interpolation. In all experiments, $h$ and $w$ are set to 8 and 32 pixels, respectively.
	
	The proposed Masked RoI has two benefits: 1) The binary mask of the target text line can eliminate the noise features caused by the background or other text lines, so as to accurately extract the features of arbitrarily-shaped text lines. 2) Masked RoI skips the spatial rectification step (\eg, STN in ASTER~\cite{shi2018aster}), reducing the time cost of feature extraction.
	
	Although our Masked RoI is somewhat analogous to RoI Masking proposed in \cite{qin2019towards}, there is a major distinction between them.
	In RoI Masking~\cite{qin2019towards}, the mask used to filter noise features is an instance-wise attention map, whose weight is soft (\ie, a floating number). This approach is not suitable for our PAN++, which cannot predict instance-wise soft masks of text lines.
	Differently, our Masked RoI removes noise by the binary mask, which can be easily generated by PA.
	Moreover, the noise weight in the binary mask is 0. Therefore, the proposed Masked RoI can remove noise more thoroughly than RoI Masking.

	\subsubsection{Recognition Head}
\begin{figure}[t]
		\centering
		\setlength{\fboxrule}{0pt}
		\fbox{\includegraphics[width=0.48\textwidth]{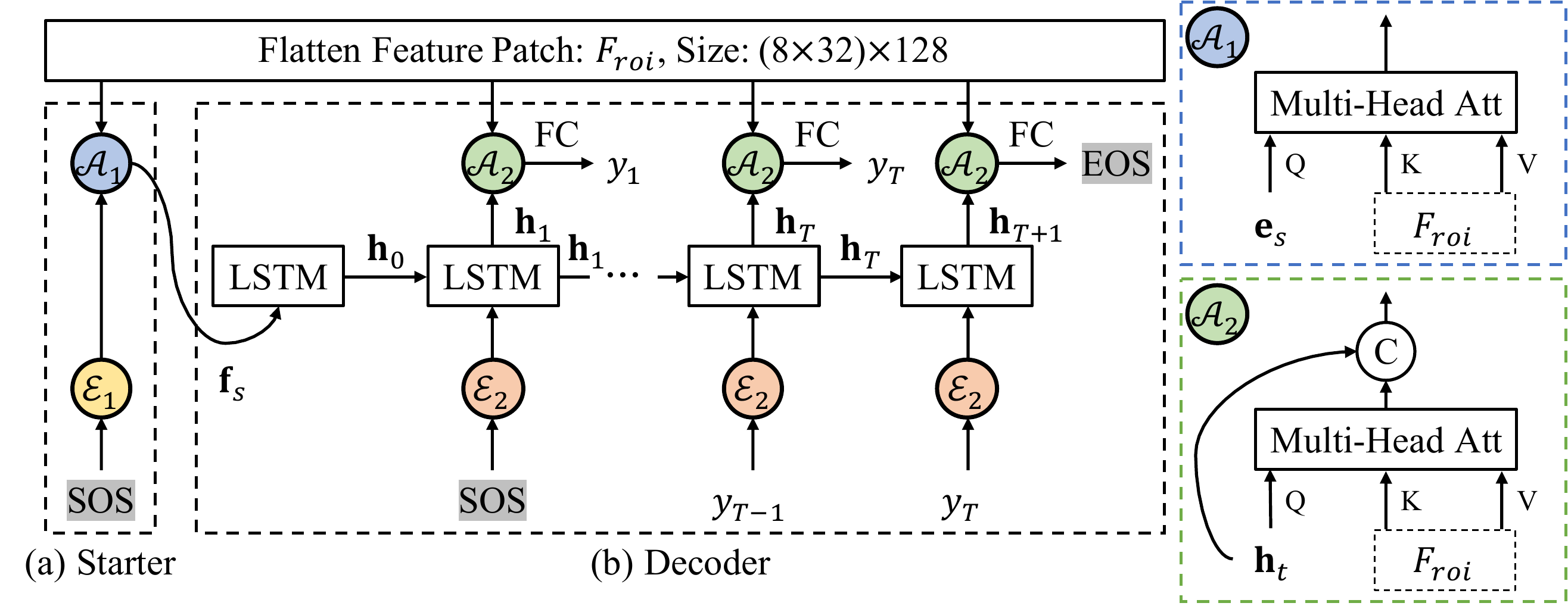}}
		\caption{\textbf{Details of the recognition head}. ``SOS'' and ``EOS'' means the start and the end of a string, respectively. $\mathcal{A}$ represents the multi-head attention layer. $\mathcal{E}$ represents the embedding layer. ${\rm C}$ means the concatenation operation in the channel dimension.
		}
		\label{fig:rec}
	\end{figure}

	Our recognition head is a seq2seq model equipped with multi-head attention~\cite{vaswani2017attention}.
	As shown in Fig.~\ref{fig:rec}, it is composed of a starter and a decoder.
	
	The starter is proposed to find the start of the string (SOS), which is not necessarily the left-most region of an arbitrarily-shaped text line.
	The starter only contains a linear
	transformation (embedding layer) $\mathcal{E}_1$ and a multi-head attention layer $\mathcal{A}_1$.
	As described in Eqn.~\eqref{eqn: fs},
	the embedding layer $\mathcal{E}_1$ transfer the SOS symbol (one-hot) to a 128-dim vector, and then this vector is fed to the multi-head attention layer $\mathcal{A}_1$ along with the flatten feature patch $F_{roi}$, producing SOS's feature vector ${\rm \textbf{f}_s}$
	, which is also the input of the decoder at the initial time step.
	\begin{equation}
		{\rm \textbf{f}}_s =\mathcal{A}_1(\mathcal{E}_1({\rm ``SOS"}), F_{roi})
		\label{eqn: fs}
	\end{equation}
	
	The decoder is made up of only two LSTM layers and one multi-head attention layer $\mathcal{A}_2$.
	At time step 0 (initial time step), the decoder takes SOS's feature vector ${\rm \textbf{f}_s}$ and zero LSTM initial states as input.
    At time step 1, we input the hidden state ${\rm \textbf{h}_0}$ of time step 0 into LSTM along with the SOS symbol, and predict the output symbol ${\rm y}_1$ of time step 1.
	After that, the output symbol of the previous step is fed into LSTM until predicting the end of the string (EOS) token. We formulate this process as follows:
	\begin{equation}
		{\rm y}_t = {\rm argmax}({\rm FC}(\mathcal{A}_2({\rm \textbf{h}}_t, F_{roi}))),~~~~1 \le t \le T,
	\end{equation}
	\begin{equation}
		{\rm \textbf{h}}_0 = {\rm LSTM}({\rm \textbf{f}}_s, {\rm \textbf{0}}),
	\end{equation}
	\begin{equation}
		{\rm \textbf{h}}_t = {\rm LSTM}(\mathcal{E}_2({\rm y}_{t-1}), {\rm \textbf{h}}_{t-1}).
	\end{equation}
	Here, \textbf{0} represents zero states. All input symbols (including SOS and EOS) are represented by one-hot vectors, followed by an embedding layer $\mathcal{E}_2$. ${\rm y}_t$ is the output symbol of the time step $t$.
	During training, the input symbols of the decoder are characters in the ground-truth sequence.
	
	Our recognition head is lightweight while maintaining good accuracy. 
	Our recognition head has no encoder, and its decoder only contains two LSTM layers and an attention layer. Moreover, our attention layer is based on multi-head attention, which can effectively fuse temporal (LSTM) features and visual (CNN) features. Besides that, the computational cost of multi-head attention is lower than the common attention module used in \cite{li2019show}.

	\subsection{Label Generation}
	As presented in Fig.~\ref{fig:arch} (f)(g), our method predicts the masks of text regions and text kernels. Therefore, the corresponding mask label is required for training.
	
	For text regions, their mask labels can be directly generated by filling ground-truth bounding boxes, where the text pixel is 1 and the non-text pixel is 0. Here, we denote the mask label of the text region as $G_{tex}$.
	
\begin{figure}[t]
		\centering
		\setlength{\fboxrule}{0pt}
		\fbox{\includegraphics[width=0.48\textwidth]{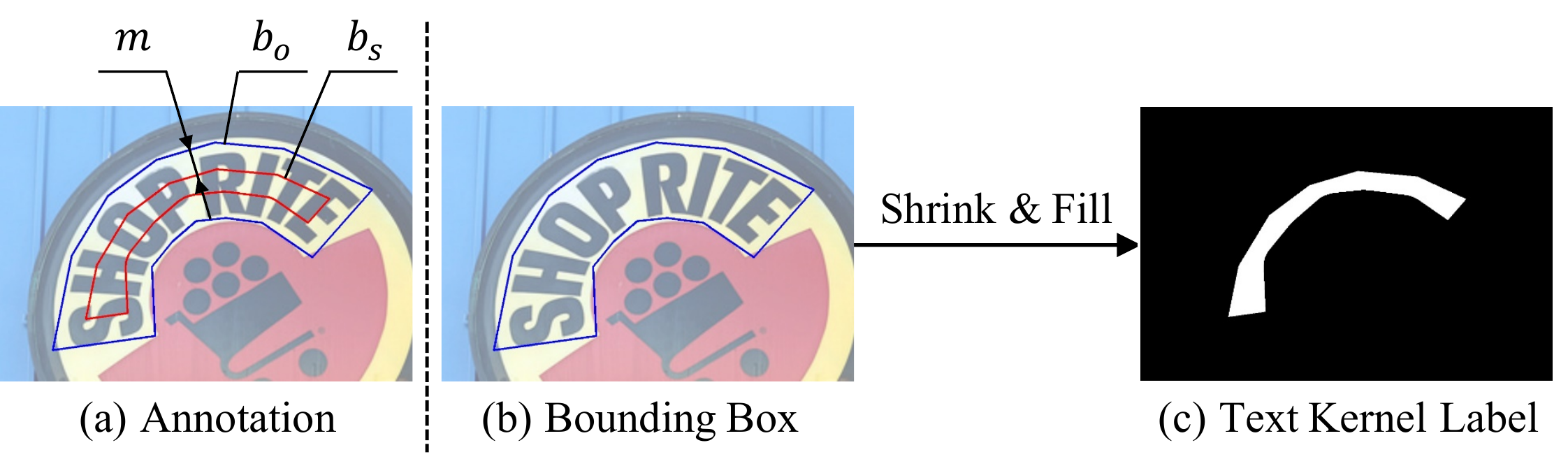}}
		\caption{\textbf{Illustration of label generation}. (a) contains the annotations of the shrinking margin $m$, the original bounding box $b_o$ and the shrunk bounding box $b_s$. (b) shows the original bounding box of a text line. (c) shows the text kernel label.}
		\label{fig:label_gen}
	\end{figure}

    For text kernels, we generate their mask labels by shrinking the original bounding boxes by a certain margin and filling them.
	As shown in Fig.~\ref{fig:label_gen}, the polygon with a blue border denotes the original bounding box of a text line.
	To generate the label of the text kernel, we firstly utilize the Vatti clipping algorithm~\cite{vatti1992generic} to shrink the original bounding box $b_o$ by $m$ pixels and get shrunk bounding box $b_k$ (see Fig.~\ref{fig:label_gen}~(a)). 
	After that, the shrunk bounding box $b_k$ is transferred into a binary mask, which is the label of the text kernel.
	For convenience, we denote the mask labels of text kernels as $G_{ker}$.
	
	During the label generation of text kernels, if we consider a shrinking rate $r\in[0, 1)$, the shrinking margin $m$ between the original bounding box $b_o$ and the shrunk bounding box $b_s$ can be calculated as:
	\begin{equation}
		m = \frac{{\rm Area}(b_o) \times (1 - r^2)}{{\rm Perimeter}(b_o)}.
		\label{eqn:d}
	\end{equation}
	Here, ${\rm Area}(\cdot)$ is the function of computing the area of a polygon. ${\rm Perimeter}(\cdot)$ is the function of computing the perimeter of a polygon. 
	In the training phase, we can control the scale of target text kernels by adjusting the shrinking rate $r$.
	
	\subsection{Loss Function}
	Our loss function can be formulated as Eqn.~\eqref{eqn:loss_tot}.
	\begin{equation}
		\mathcal{L} = \mathcal{L}_{det} + \mathcal{L}_{rec},
		\label{eqn:loss_tot}
	\end{equation}
	where $\mathcal{L}_{det}$ and $\mathcal{L}_{rec}$ are the loss functions for text detection and recognition, respectively. 
	
	Specifically, the loss function $\mathcal{L}_{det}$ of text detection can be written as:
	\begin{equation}
		\mathcal{L}_{det} =  \mathcal{L}_{tex} + \alpha \mathcal{L}_{ker} + \beta (\mathcal{L}_{agg} + \mathcal{L}_{dis}),
		\label{eqn:loss_det}
	\end{equation}
	where $\mathcal{L}_{tex}$ is the loss for text region segmentation. $\mathcal{L}_{ker}$ is the loss for text kernel segmentation.
	$\alpha$ and $\beta$ are used to balance the importance of $\mathcal{L}_{tex}$, $\mathcal{L}_{ker}$, $\mathcal{L}_{agg}$ and $\mathcal{L}_{dis}$. We set $\alpha$ to 0.5 and $\beta$ 0.25 in all experiments.
	
	Considering the extreme imbalance of text and non-text pixels, we follow~\cite{psenet,wang2019efficient} and adopt dice loss~\cite{milletari2016v} to optimize the segmentation result of text regions $P_{tex}$ and of text kernels $P_{ker}$.
	Therefore, $\mathcal{L}_{tex}$ and $\mathcal{L}_{ker}$ are formulated as:
	\begin{equation}
		\mathcal{L}_{tex} = 1 - \frac{2 \sum_{i} P_{tex}(i)G_{tex}(i)}{\sum_i P_{tex}(i)^2 + \sum_i G_{tex}(i)^2},
		\label{eqn:loss_tex}
	\end{equation}
	\begin{equation}
		\mathcal{L}_{ker} = 1 - \frac{2 \sum_{i} P_{ker}(i)G_{ker}(i)}{\sum_i P_{ker}(i)^2 + \sum_i G_{ker}(i)^2}.
		\label{eqn:loss_ker}
	\end{equation}
	Here, $P_{tex}(i)$ and $G_{tex}(i)$ refer to the value of the $i$th pixel in the segmentation result and the ground truth of the text regions, respectively. Similarly, $P_{ker}(i)$ and $G_{ker}(i)$ means the $i$th pixel value in the prediction and the ground truth of text kernels, respectively.
	We also adopt Online Hard Example Mining (OHEM)~\cite{shrivastava2016training} to ignore simple non-text pixels when calculating ${L}_{tex}$. Note that, we only take text pixels into consideration when calculating $\mathcal{L}_{ker}$, $\mathcal{L}_{agg}$ and $\mathcal{L}_{dis}$.
	
	The loss function $\mathcal{L}_{rec}$ of text recognition is:
	\begin{equation}
		\mathcal{L}_{rec} = \frac{1}{\left|w\right|} \sum_{i=0}^{\left|w\right|} {\rm CrossEntropy}(y_i, w_i),
		\label{eqn:loss_rec}
	\end{equation}
	where $w$ is the ground-truth transcription (text content) containing EOS symbol. $\left|w\right|$ denotes the number of characters in the transcription. $w_i$ is the $i$th character in the transcription.
	
	\section{Experiment}
	To validate the effectiveness of the proposed PAN++, we evaluate it on two challenging tasks: 1) text detection and 2) end-to-end text spotting.
	\begin{table*}[t]
		\centering
		\renewcommand\arraystretch{1}
		\caption{\textbf{Text detection results on Total-Text and CTW1500}. ``$*$'' indicates the results from \cite{textsnake}. ``$\dagger$'' indicates the results from \cite{Liu2017Detecting}. 
		\revise{``Scale'' denotes the scale of the testing image, where ``\emph{L}:'' means the longer side is fixed, and ``\emph{S}:'' means the shorter side is fixed.}
		Note that, we use 
		a 
		1080Ti GPU to test the inference speed of PAN++ (detection only) for fair comparisons.
		Precision (\%), Recall (\%) and F-measure (\%)
		are reported here.
		}
		\setlength{\tabcolsep}{2.0mm}
		\newcommand{\tabincell}[2]{\begin{tabular}{@{}#1@{}}#2\end{tabular}}
\begin{tabular}{r|c|c|c|c|c|c|c|c|c|c|c}
	\whline 
	\multirow{2}{*}{Method} & \multirow{2}{*}{Scale} &
	\multirow{2}{*}{Backbone} &
	\multirow{2}{*}{\tabincell{c}{External\\Dataset}} & 
	\multicolumn{4}{c|}{Total-Text} &
	\multicolumn{4}{c}{CTW1500}
	 \\
	\cline{5-12}
	& & & & Precision & Recall & F-measure & FPS & Precision & Recall & F-measure & FPS\\
	\hline
	\hline
	CTPN~\cite{tian2016detecting} & \emph{S}: 600 & VGG16 & $\times$  & - & - & - & - & 60.4$^\dagger$ & 53.8$^\dagger$ & 56.9$^\dagger$ & 7.1$^\dagger$ \\
	\hline
	SegLink~\cite{shi2017detecting}& 768$\times$1280 & VGG16 & $\times$ & 30.3$^*$ & 23.8$^*$ & 26.7$^*$ & - & 42.3$^\dagger$ & 40.0$^\dagger$ & 40.8$^\dagger$ & 10.7$^\dagger$ \\
	\hline
	EAST~\cite{zhou2017east}& 720$\times$1280 & VGG16 & $\times$ & 50.0$^*$ & 36.2$^*$ & 42.0$^*$ & - & 78.7$^\dagger$ & 49.1$^\dagger$ & 60.4$^\dagger$ & 21.2$^\dagger$ \\
	\hline
	DeconvNet~\cite{totaltext} & 224$\times$224 & VGG16 & $\times$ & 33.0 & 40.0 & 36.0 & - & - & - & - & -  \\
	\hline
	CTD+TLOC~\cite{Liu2017Detecting}& \emph{S}: 600 & ResNet50 & $\times$ & - & - & - & - & 77.4 & 69.8 & 73.4 & 13.3 \\
	\hline
	\multirow{3}{*}{PAN++ (ours)}
	& \emph{S}: 320 & ResNet18 & $\times$ & 82.6 & 72.9 & 77.4 & \textbf{84.9} & 82.9 & 76.3 & 79.5 & \textbf{84.2} \\
	& \emph{S}: 512 & ResNet18 & $\times$ & 87.1 & 79.0 & 82.8 & 53.1 & 85.2 & 80.3 & 82.7 & 52.7 \\
	& \emph{S}: 640 & ResNet18 & $\times$ & 89.2 & 80.3 & 84.5 & 38.3 & 85.2 & 81.1 & 83.1 & 36.0 \\
	\hline
	\hline
	TextSnake~\cite{textsnake}& 512$\times$512 & VGG16 & SynthText & 82.7 & 74.5 & 78.4 & 12.4 & 67.9 & 85.3 & 75.6 & - \\
	\hline
	TextField~\cite{xu2019textfield} & 768$\times$768 & VGG16 & SynthText & 81.2 & 79.9 & 80.6 & - & 83.0 & 79.8 & 81.4 & - \\
	\hline
	CRAFT~\cite{baek2019character} & \emph{L}: 1280 & VGG16 & SynthText & 87.6 & 79.9 & 83.6 & 4.8 & 86.0 & 81.1 & 83.5 & 7.6 \\
	\hline
	LOMO~\cite{zhang2019look} & \emph{S}: 512 & ResNet50 & SynthText & 88.6 & 75.7 & 81.6 & 4.4 & 89.2 & 69.6 & 78.4 & - \\
	\hline
	SPCNet~\cite{spcnet}& \emph{S}: 800 & ResNet50 & IC17-MLT & 83.0 & 82.8 & 82.9 & 4.6 & - & - & - & -  \\
	\hline
	\multirow{3}{*}{PAN++ (ours)}
	& \emph{S}: 320 & ResNet18 & SynthText & 85.6 & 75.1 & 80.0 & \textbf{84.9} & 84.5 & 76.5 & 80.3 & \textbf{84.2} \\
	& \emph{S}: 512 & ResNet18 & SynthText & 88.4 & 80.5 & 84.2 & 53.1 & 86.9 & 80.4 & 83.5 & 52.7 \\
	& \emph{S}: 640 & ResNet18 & SynthText & \textbf{89.9} & \textbf{81.0}  & \textbf{85.3} & 38.3 & \textbf{87.1} & 81.1 & \textbf{84.0} & 36.0 \\
	\whline 
\end{tabular}
		\label{tab:ct_det}
	\end{table*}
	\subsection{Datasets}
	
	\subsubsection{Curved Text Datasets}
	\noindent\textbf{Total-Text}~\cite{totaltext} is a dataset for arbitrarily-shaped text detection and spotting, containing horizontal, multi-oriented, and curved text lines. This dataset consists of 1,255 training images and 300 testing images, all of which are annotated with polygons and transcriptions at the word level.
	
	\noindent\textbf{CTW1500}~\cite{Liu2017Detecting} is a widely used dataset for arbitrarily-shaped text detection. It has 1,000 training images and 500 testing images. Different from the Total-Text dataset, the images in this dataset are labeled at the text line level. The text lines in the CTW1500 dataset are long and annotated by 14-points polygons.
	
	\subsubsection{Straight Text Datasets}
	\textbf{ICDAR 2015} (IC15)~\cite{karatzas2015icdar} is a commonly used dataset for end-to-end text detection and recognition. It contains a total of 1,500 images, 1,000 of which are used for training and the remaining are for testing. In this dataset, text lines are annotated by quadrangles and transcriptions at the word level.
	
	\noindent\textbf{MSRA-TD500}~\cite{yao2012detecting} includes 300 training images and 200 test images with text line level annotations. It contains multi-lingual, multi-oriented, and long text lines, and all of them are labeled at the text line level. Because its training set is small, we follow the previous works \cite{zhou2017east,lyu2018multi,textsnake} to include the 400 images of HUST-TR400~\cite{yao2014unified} as training data.
	
	\noindent\revise{\textbf{RCTW-17}~\cite{shi2017icdar2017} is a competition on reading Chinese Text in images. It provide a large-scale dataset that consists of 12,000 images, where 8,346 for training and the rest for testing. Every image in this dataset is annotated with text line quadrilaterals and text transcripts.}
	
	\subsubsection{Datasets for Pre-training or Jointly Training}
	\textbf{SynthText}~\cite{synthtext} is a large-scale synthetically generated dataset containing 800K synthetic images. There are a large number of multi-oriented text lines, which are annotated at word-level and character-level rotated bounding boxes, as well as corresponding transcriptions. Following \cite{shi2017detecting,lyu2018multi,textsnake}, this dataset is used for pre-training or jointly training models of PAN++.

	\noindent\textbf{COCO-Text}~\cite{veit2016coco} contains 63,686 images and three versions of the annotations (V1.1, V1.4, and V2.0). In V2.0, there are 239,506 annotated text lines, which are labeled with word-level polygons and transcriptions. This dataset is usually used for pre-training or jointly training in previous text spotters~\cite{qin2019towards,liu2020abcnet}.
	
	\noindent\textbf{ICDAR 2017 MLT} (IC17-MLT)~\cite{nayef2017icdar2017} is a multi-language scene text dataset. It consists of 7,200 training images, 1,800 validation images, and 9,000 test images.  The dataset is composed of natural scene images with nine languages, and these images are labeled with word-level quadrangles and transcriptions. Following \cite{qin2019towards,liu2020abcnet}, we use the English samples in the training set to pre-train or jointly train our text spotting models.
	
	\subsection{Experiments on Curved Text Datasets}
	\subsubsection{Experiment Settings}
	\label{sec:ct_setting}
	To test the performance of PAN++ on arbitrarily-shaped text, we conduct experiments on Total-Text and CTW1500 and compare it with previous start-of-the-art methods. In these experiments, we use the ResNet~\cite{he2016identity} pre-trained on ImageNet~\cite{deng2009imagenet} as the backbone network of our method. In addition, we set the dimension of the instance vector to 4, the negative-positive ratio of OHEM to 3, the shrinking rate $r$ of the text kernel to 0.7, and the distance threshold $d$ of PA to 3.
	
	Following the common practices~\cite{zhou2017east,psenet,masktextspotter,liu2020abcnet}, we ignore the blurred text regions labeled as ``DO NOT CARE'' during training, and apply random scale, random horizontal flip, random rotation, and random crop on training images. 
	All models are optimized by using ADAM~\cite{kingma2014adam} optimizer with a batch size of 16 on 4 GPUs. The initial learning rate is set to $1\times10^{-3}$. 
	Similar to \cite{zhao2017pyramid}, we use the ``poly'' learning rate strategy in which the initial rate is multiplied by $(1 - \frac{\rm iter}{\rm max\ iter})^{\rm power}$, and the $\rm power$ is set to 0.9 in all experiments.
	In the testing phase, we resize 
	the input image
	to different scales and report the performance on text detection and end-to-end text spotting tasks. All results are tested with a batch size of 1 on a V100 GPU and a 2.20GHz CPU in a single thread unless otherwise stated.
	
	In the text detection task, to make fair comparisons, we train PAN++ (without the recognition head) with two widely used strategies as follows: 
	1) following \cite{rrpn,zhou2017east,wang2019efficient} to train models without external text datasets; 2) following \cite{shi2017detecting,wang2019efficient,psenet} to fine-tune models pre-trained on external text dataset (\eg, SynthText and IC17-MLT).
	In the first strategy, we train our method for 36K iterations. In the second one, we first pre-train models on the external text dataset for 50K iterations and then fine-tune the pre-trained models for 36K iterations.
	
	In the text spotting task, the training strategies used in previous works~\cite{liu2018fots,liu2020abcnet,masktextspotter,qin2019towards} can also be divided into two categories: 1) fine-tuning models pre-trained on external text datasets, and 2) jointly training models on multiple text datasets.
	For fair comparisons, we train our end-to-end text spotting models using the two strategies respectively.
	Specifically, when using the fine-tuning strategy, we first pre-train models on the training images in the SynthText, COCO-Text, and IC17-MLT datasets for 150K iterations, and then we fine-tune the pre-trained models on Total-Text or CTW1500 for 7K iterations.
	For the jointly training strategy, we train our models for 150K iterations on the joint dataset that includes all training samples in SynthText, COCO-Text, IC17-MLT, Total-Text and IC15.
	
	\subsubsection{Curved Text Detection Results}
	\revise{
	As shown in Table \ref{tab:ct_det}, 
	without pre-training on external text datasets,
	PAN++ with the shorter side being 320 pixels yields an F-measure of 79.5\% on CTW1500, surpassing most of the counterparts, including some methods (\eg, TextSnake~\cite{textsnake}) pre-trained on SynthText (79.5\% \vs 78.4\%).
	Moreover, our inference speed is 8 times faster than TextSnake and 4 times faster than EAST~\cite{zhou2017east} (the fastest method before).
	It notable that when fine-tuning the model pre-trained on SynthText, PAN++ (\emph{S}: 640) achieves the best F-measure of 84.0\%, outperforming all counterparts while still keeping the fastest inference speed (36 FPS).
	}
	
	\revise{
    Similar results are also on the Total-Text dataset.
	When the shorter side is 320 pixels, the inference speed of our method reaches 84.9 FPS, which is at least 6 times faster than previous methods, while the F-measure is still very competitive (80.0\%).
	The best F-measure of our method is 85.3\%, surpassing current state-of-the-art methods over 1.7 points. Meanwhile, its speed is close to 40 FPS, which is still 3 times faster than the second-fastest TextSnake~\cite{textsnake}.
	}

	These results prove the superiority of the proposed PAN++ (detection only) in arbitrarily-shaped text detection, in terms of accuracy and speed.
    We present some qualitative curved text detection results in Fig.~\ref{fig:det_res}~(a)(b), which demonstrates that the detection components of PAN++ can elegantly locate text lines with complex shapes.
	
	\subsubsection{Curved Text Spotting Results}
	\begin{table*}[t]
		\centering
		\renewcommand\arraystretch{1}
		\caption{\textbf{End-to-end text spotting results on the Total-Text dataset}. ``None'' means no lexicon is used. ``Full'' means the lexicon contains all words in test set is used. \revise{``Scale'' denotes the scale of the testing image, where ``\emph{L}:'' means the longer side is fixed, and ``\emph{S}:'' means the shorter side is fixed, and ``\emph{MS}'' means multi-scale testing.}
		F-measure (\%) is reported here.
		}
		\setlength{\tabcolsep}{2.0mm}
		\begin{tabular}{r|c|c|c|l|c|c|c}
	\whline
	\multirow{2}{*}{Method} & \multirow{2}{*}{Scale} & 
	\multirow{2}{*}{Backbone} &
	\multirow{2}{*}{Training Strategy} &
	\multirow{2}{*}{External Dataset} &
	\multicolumn{3}{c}{Total-Text}
	 \\
	\cline{6-8}
	& & & & & None & Full & FPS\\
	\hline
	\hline
	TextNet~\cite{sun2018textnet} & \emph{L}: 920 & ResNet50 & Finetune & SynthText  & 54.0 & - & 2.7 \\
	\hline
	CharNet~\cite{xing2019charnet} & \emph{L}: 2280 & ResNet50 & Finetune & SynthText, IC15, IC17-MLT  & 66.2 & - & 1.2 \\
	\hline
	TextDragon~\cite{feng2019textdragon} & - & VGG16 &Finetune & SynthText, IC15  & 48.8 & 74.8 & - \\
	\hline
	ABCNet~\cite{liu2020abcnet} & \emph{S}: 800 & ResNet50 &Finetune& SynthText, COCO-Text, IC19-MLT & 64.2 & 75.7 & 17.9 \\
	\hline
	\multirow{3}{*}{PAN++ (ours)}
	& \emph{S}: 512 & ResNet18 &Finetune& SynthText, COCO-Text, IC17-MLT & 63.5 & 74.0 & \textbf{29.2} \\
	& \emph{S}: 640 & ResNet18 &Finetune& SynthText, COCO-Text, IC17-MLT & 66.4 & 77.3 & 24.1 \\
	& \emph{S}: 736 & ResNet18 &Finetune& SynthText, COCO-Text, IC17-MLT & \textbf{67.1} & \textbf{77.4} & 21.1\\
	\hline
	\hline
	TextBoxes~\cite{liao2017textboxes} & \emph{MS} & ResNet50 &Jointly Train& SynthText, IC13, IC15 &  36.3 & 48.9 & 1.4 \\
	\hline
	Mask TextSpotter~\cite{masktextspotter} & \emph{S}: 1000  & ResNet50 &Jointly Train& SynthText, IC13, IC15 & 52.9 & 71.8 & 4.8 \\
	Mask TextSpotter v2~\cite{liao2019mask}& \emph{S}: 1000 & ResNet50 &Jointly Train& SynthText, IC13, IC15, SCUT & 65.3 & 77.4 & - \\
	\hline
	Qin \emph{et al.}~\cite{qin2019towards} & \emph{S}: 900 & ResNet50 &Jointly Train& SynthText, IC15, COCO-Text, IC17-MLT, Private Data &  67.8 & - & 4.8 \\
	\hline
	\multirow{3}{*}{PAN++ (ours)}
	& \emph{S}: 512 & ResNet18 &Jointly Train& SynthText, IC15, COCO-Text, IC17-MLT & 64.9 & 75.7 & \textbf{29.2} \\
	& \emph{S}: 640 & ResNet18 &Jointly Train& SynthText, IC15, COCO-Text, IC17-MLT & 66.7 & 77.5 & 24.1 \\
	& \emph{S}: 736 & ResNet18 &Jointly Train& SynthText, IC15, COCO-Text, IC17-MLT & \textbf{68.6} & \textbf{78.6} & 21.1 \\
	\whline
	
\end{tabular}
		\label{tab:ct_reg}
	\end{table*}
	
	\revise{
	End-to-end text spotting results on the Total-Text dataset are reported in Table \ref{tab:ct_reg}.
	Compared with the previous fastest ABCNet~\cite{liu2020abcnet}, our method with shorter side being 640 pixels runs 1.3 times faster, while our end-to-end text spotting F-measure is 2.2 points higher (66.4\%  \vs 64.2\%).
	When adopting a larger input scale or jointly training strategy, the performance of our method can be further improved.
	With low-resolution input (the shorter side being 512 pixels), the inference speed of our method reaches 29.2, at least 11 FPS faster than previous methods.
    At the same time, its end-to-end text spotting F-measure is 64.9\%, which is higher than most counterparts.
	It is notable that PAN++ achieves the best end-to-end text spotting F-measure of 68.6\%, which is much better than the second-best method (Qin \emph{et al}.~\cite{qin2019towards}) and our speed is 4 times faster (21.1 FPS \vs 4.8 FPS).
	}
	
	These results indicate that the proposed method achieves state-of-the-art performance in the text spotting task, significantly outperforming existing methods, especially in the inference speed.
	Some qualitative results of end-to-end curved text spotting in shown in Fig.~\ref{fig:rec_res} (a).

	\subsection{Experiments on Straight Text Datasets}
	\subsubsection{Experiment Settings}
	\begin{table*}[t]
		\centering
		\renewcommand\arraystretch{1}
		\caption{\textbf{Text detection results on IC15 and MSRA-TD500}. \revise{``Scale'' denotes the scale of the testing image, where ``\emph{L}:'' means the longer side is fixed, and ``\emph{S}:'' means the shorter side is fixed, and ``\emph{MS}'' means multi-scale testing.} 
		Note that, we use 1080Ti GPU to test the inference speed of PAN++ (detection only) for fair comparisons.}
		\setlength{\tabcolsep}{2.0mm}
		\newcommand{\tabincell}[2]{\begin{tabular}{@{}#1@{}}#2\end{tabular}}
\begin{tabular}{r|c|c|c|c|c|c|c|c|c|c|c}
	\whline
	\multirow{2}{*}{Method} & \multirow{2}{*}{Scale} &
	\multirow{2}{*}{Backbone} &
	\multirow{2}{*}{\tabincell{c}{External\\Dataset}} & 
	\multicolumn{4}{c|}{IC15} &
	\multicolumn{4}{c}{MSRA-TD500}
	 \\
	\cline{5-12}
	& & & & Precision & Recall & F-measure & FPS & Precision & Recall & F-measure & FPS\\
	\hline
	\hline
	CTPN~\cite{tian2016detecting} & \emph{S}: 600 & VGG16 & $\times$ & 74.2 & 51.6 & 60.9 & 7.1 & - & - & - & - \\
	\hline
	RRPN~\cite{rrpn} & $L$: 1000  & VGG16 & $\times$  & 82.0 & 73.0 & 77.0 & - & 82.0 & 68.0 & 74.0 & - \\
	\hline
	EAST~\cite{zhou2017east} & 720$\times$1280 & VGG16  & $\times$ & 83.6 & 73.5 & 78.2 & 13.2 & 87.3 & 67.4 & 76.1 & - \\
	\hline
	PixelLink~\cite{PixelLink} & 768$\times$1280 & VGG16 & $\times$  & 82.9 & 81.7 & 82.3 & 7.3 & 81.1 & 73.0 & 76.8 & 3.0\\
	\hline
	DeepReg~\cite{deepreg} & \emph{MS} & VGG16 & $\times$ & 82.0 & 80.0 & 81.0 & - & 77.0 & 70.0 &74.0 & 1.1\\
	\hline
	\multirow{2}{*}{PAN++ (ours)}
	& \emph{S}: 736 & ResNet18 & $\times$ & 85.5 & 77.2 & 81.2 & \textbf{28.2} & 81.6 & 80.3 & 80.9 & \textbf{32.5} \\
	& \emph{S}: 896 & ResNet18 & $\times$ & 86.7 & 78.4 & 82.3 & 19.2  & 82.4 & 82.1 & 82.2 & 22.6\\
	\hline
	\hline
	SegLink~\cite{shi2017detecting} & 768$\times$1280 & VGG16 & SynthText & 73.1 & 76.8 & 75.0 & - & 86.0 & 70.0 & 77.0 & 8.9  \\
	\hline
	SSTD~\cite{he2017single} & 704$\times$704 & VGG16 & Private Data & 80.2 & 73.9 & 76.9 & 7.7 & - & - & - & - \\
	\hline
	RRD~\cite{rrd} & 1024$\times$1024 & VGG16 & SynthText & 85.6 & 79.0 & 82.2 & 6.5 & 87.0 & 73.0 & 79.0 & 10 \\
	\hline
	MCN~\cite{mcn} & 512$\times$512 & VGG16 & SynthText & 72.0 & 80.0 & 76.0 & - & 88.0 & 79.0 & 83.0 &-\\
	\hline
	TextSnake~\cite{textsnake} & 768$\times$1280 & VGG16 & SynthText & 84.9 & 80.4 & 82.6 & 1.1 & 83.2 & 73.9 & 78.3 & 1.1 \\
	\hline
	LOMO~\cite{zhang2019look} & \emph{L}: 1536 & ResNet50 & SynthText & 91.3 & 83.5 & 87.2 & - & - & - & - & - \\
	\hline
	SPCNet~\cite{spcnet} & \emph{S}: 800 & ResNet50 & IC17-MLT & 88.7 & \textbf{85.8} & 87.2 & 4.6 & - & - & - \\
	\hline
	\multirow{3}{*}{PAN++ (ours)}
	& \emph{S}: 736 & ResNet18 & SynthText & 85.9 & 80.4 & 83.1 & \textbf{28.2} & 85.3 & 84.0 & 84.7 & \textbf{32.5} \\
	& \emph{S}: 896 & ResNet18 & SynthText & 88.7 & 80.7 & 84.5 & 19.2  & 89.6 & 86.3 & 87.9 & 22.6\\
	& \emph{S}: 896 & ResNet50 & IC17-MLT & \textbf{91.4} & 83.9 & \textbf{87.5} & 12.6 & \textbf{91.4} & \textbf{85.6} & \textbf{88.4} & 14.3 \\
	\whline
\end{tabular}
		\label{tab:st_det}
	\end{table*}
	
	We also evaluate the performance of PAN++ on straight text datasets, including IC15 and MSRA-TD500. IC15 is a representative benchmark for the text detection task and the text spotting task, and MSRA-TD500 is a dataset annotated at the text line level and often used to benchmark long text detection.
	During training, the shrinking rate $r$ on IC15 and MSRA-TD500 are set to 0.5 and 0.7, respectively.
	Other model settings, data augmentations, and training settings are the same as those in Sec. \ref{sec:ct_setting}. 
	In the testing phase, we rescale the shorter side of the input image to 736 or 896 pixels for fair comparisons with previous methods.
	
	\subsubsection{Straight Text Detection Results}
	Straight text detection results are presented in Table \ref{tab:st_det}. On the IC15 dataset, PAN++ (the shorter side being 736 pixels) yields an F-measure of 81.2\% at 28.2 FPS without pre-training on external text dataset. Compared with the previous fastest method EAST~\cite{zhou2017east}, our method surpasses EAST by 3.0 points in F-measure, while our method runs 2 times faster.
	Fine-tuning on the SynthText dataset can further increase the F-measure of our method to 83.1\%, which is 0.5 points higher than TextSnake~\cite{textsnake}, but our method can run 25 times faster.
	\revise{
	Because IC15 contains many small texts, previous methods~\cite{spcnet,rrd,zhang2019look} usually employ high-resolution input and a heavy backbone to ensure detection accuracy.
	With these setting, the best F-measure of our method reaches 87.5\%, which is higher than all counterparts. Even in this case, our inference speed is still the fastest (12.6 FPS).
	}
	
	On the MSRA-TD500 dataset, 
	our method yields F-measures of 80.9\% and 84.7\% when using the external text dataset or not. Compared with previous state-of-the-art methods, our method enjoys higher accuracy and faster inference speed. Notably, the best F-measure of our method is 88.4\%, which is at least 5.4 points higher than previous methods. This result indicates that our method is also robust for long straight text detection.
	
\begin{figure*}[t]
		\begin{center}
			\setlength{\fboxrule}{0pt}
			\fbox{\includegraphics[width=0.9\textwidth]{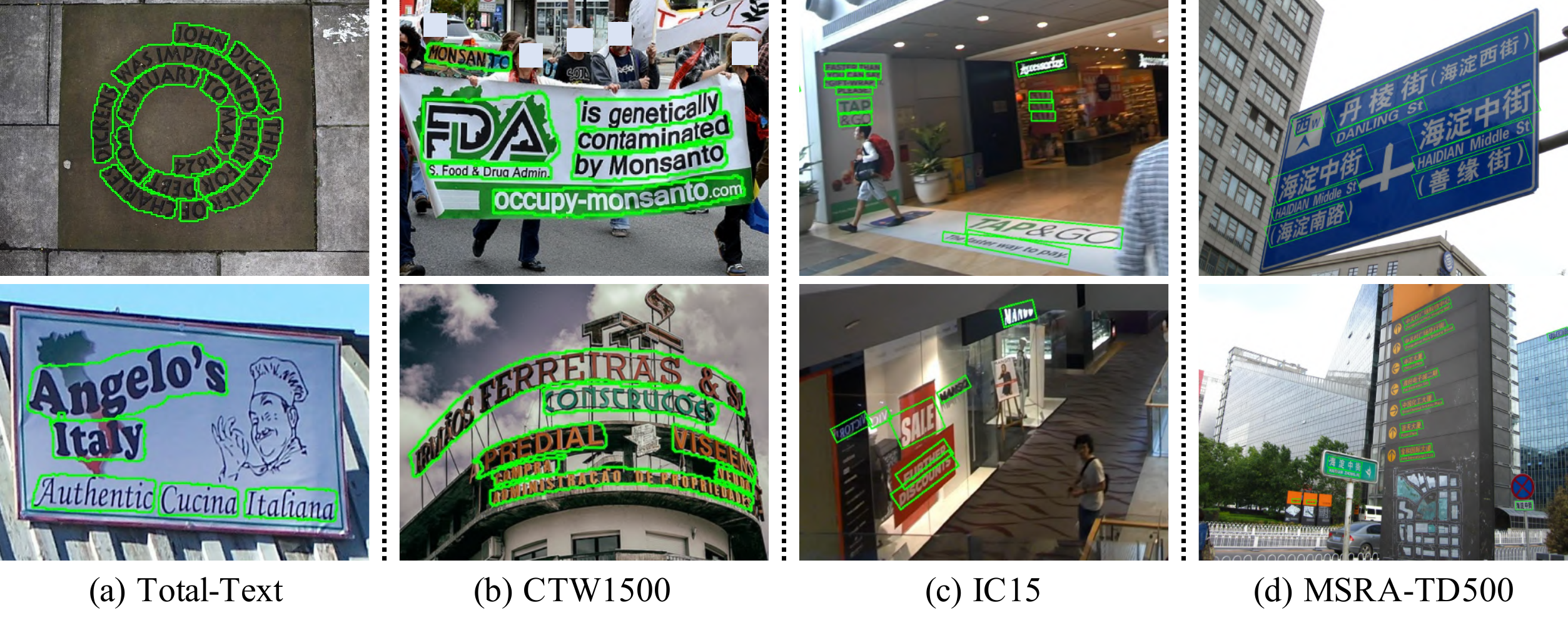}}
			\caption{\textbf{Qualitative text detection results of PAN++ on Total-Text~\cite{totaltext}, CTW1500~\cite{Liu2017Detecting}, IC15~\cite{karatzas2015icdar} and MSRA-TD500~\cite{msra}}. Our method works well on various complex natural scenes, for example, arbitrary shapes, multiple languages, extreme illumination.}

			\label{fig:det_res}
		\end{center}
	\end{figure*}

	In conclusion, the proposed PAN++ (detection only) can also accurately detect straight text lines.
	Some qualitative results of straight text detection are shown in Fig.~\ref{fig:det_res} (c)(d).
	
	\subsubsection{Straight Text Spotting Results}
	\begin{table*}[t]
		\centering
		\renewcommand\arraystretch{1}
		\caption{\textbf{End-to-end text spotting results on the IC15 dataset}. ``S'' (strong) means 100 words, including the ground truth, are given for each image. For ``W'' (weak), a lexicon includes all words that appear in the test set is provided. For ``G'' (generic), a generic lexicon with 90k words is given. ``N'' (none) means that no lexicon is used. \revise{``Scale'' denotes the scale of the testing image, where ``\emph{L}:'' means the longer side is fixed, and ``\emph{S}:'' means the shorter side is fixed, and ``\emph{MS}'' means multi-scale testing.}}
		\setlength{\tabcolsep}{0.75mm}
		\begin{tabular}{r|c|c|c|l|c|c|c|c|c}
	\whline
	\multirow{2}{*}{Method} & \multirow{2}{*}{Scale} & 
	\multirow{2}{*}{Backbone} &
	\multirow{2}{*}{Training Strategy} &
	\multirow{2}{*}{External Dataset} &
	\multicolumn{5}{c}{IC15}
	 \\
	\cline{6-10}
	& & & & & S & W & G & N & FPS\\
	\hline
	\hline
	Stradvision~\cite{karatzas2015icdar} & - & - & - & - & 43.7 & - & - & - & - \\
	\hline
	MCLAB~\cite{shi2017detecting,crnn} & - & - & - &  & 67.9 & - & - & - & - \\
	\hline
	He \emph{et al.}~\cite{he2018end} & - & PVA & Finetune & SynthText & 82 & 77  & 63 & - & - \\
	\hline
	FOTS~\cite{liu2018fots} & \emph{L}: 2240 & ResNet50 & Finetune & IC17-MLT  & 81.1 & 75.9 & 60.8 & - & 7.5 \\
	\hline
	CharNet~\cite{xing2019charnet} & \emph{L}: 2280 & ResNet50 & Finetune & SynthText, IC17-MLT, Total-Text  & 82.4 & 78.9 & 67.6 & 62.7 & 1.2 \\
	\hline
	TextNet~\cite{sun2018textnet} & \emph{MS} & ResNet50 & Finetune & SynthText & 78.7 & 74.9 & 60.5 & -  \\
	\hline
	TextDragon~\cite{feng2019textdragon} & - & VGG16 & Finetune & SynthText, Total-Text  & 82.5 & 78.3 & 65.2 & - & - \\
	\hline
	\multirow{3}{*}{PAN++ (ours)}
	& \emph{S}: 736 & ResNet18 &Finetune& SynthText, COCO-Text, IC17-MLT & 79.4 & 74.9 & 65.6 & 64.7 & \textbf{24.5} \\
	& \emph{S}: 896 & ResNet18 &Finetune& SynthText, COCO-Text, IC17-MLT & 80.6 & 76.0 & 66.4 & 65.7 & 18.6 \\
	& \emph{S}: 896 & ResNet50 &Finetune& SynthText, COCO-Text, IC17-MLT & 82.5 & 77.4 & 68.7 & 67.6 & 13.8  \\
	\hline
	\hline
	Deep TextSpotter~\cite{busta2017deep} & 608$\times$608 & Darknet19 & Jointly Train& SynthText, IC13 & 54.0 & 51.0 & 47.0 & - & 9.0 \\
	\hline
	Mask TextSpotter~\cite{masktextspotter} & \emph{S}: 1000 & ResNet50 &Jointly Train& SynthText, IC13, Total-Text & 77.3 & 69.9 & 60.3 & - & 4.8 \\
	Mask TextSpotter v2~\cite{liao2019mask} & \emph{S}: 720 & ResNet50 &Jointly Train& SynthText, IC13, Total-Text, SCUT & 74.2 & 69.2 & 63.5 & - & 3.8 \\
	\hline
	Qin \emph{et al.}~\cite{qin2019towards} & \emph{S}: 900 & ResNet50 &Jointly Train& SynthText, COCO-Text, IC17-MLT, Total-Text, Private Data & \textbf{83.4} & \textbf{79.9} & 68.0 & - & 4.8 \\
	\hline
	\multirow{3}{*}{PAN++ (ours)}
	& \emph{S}: 736 & ResNet18 &Jointly Train& SynthText, COCO-Text, IC17-MLT, Total-Text & 80.5  & 75.8 & 66.2 & 65.6 & \textbf{24.5} \\
	& \emph{S}: 896 & ResNet18 &Jointly Train& SynthText, COCO-Text, IC17-MLT, Total-Text & 81.3 & 76.4 & 67.4 & 66.6 & 18.6 \\
	& \emph{S}: 896 & ResNet50 &Jointly Train& SynthText, COCO-Text, IC17-MLT, Total-Text & 82.7 & 78.2 & \textbf{69.2} & \textbf{68.0}  & 13.8 \\
	\whline
	
\end{tabular}
		\label{tab:st_reg}
	\end{table*}
	
	\revise{
	Table \ref{tab:st_reg} summarizes end-to-end text spotting results on the IC15 dataset. 
	Note that, the F-measure with a generic lexicon (denoted as ``G'') is the most important metric in the IC15 dataset.
	When the shorter side is 736 pixels, the speed of our method reaches 24.5 FPS, which is 3 to 7 times faster than previous methods. Meanwhile, it still has a competitive ``G'' F-measure of 66.2, surpassing most counterparts including Mask TextSpotter V2~\cite{liao2019mask} (66.2\% \vs 63.5\%).
	With ResNet50 as the backbone, our method achieves the best ``G'' F-measure of 69.2\%, which is 1.2 points higher than the previous best text spotter (Qin \etal\cite{qin2019towards}). At the same time, our inference speed is still the fastest (13.8 FPS).
	}
	
\begin{figure*}[t]
		\centering
		\setlength{\fboxrule}{0pt}
		\fbox{\includegraphics[width=1.0\textwidth]{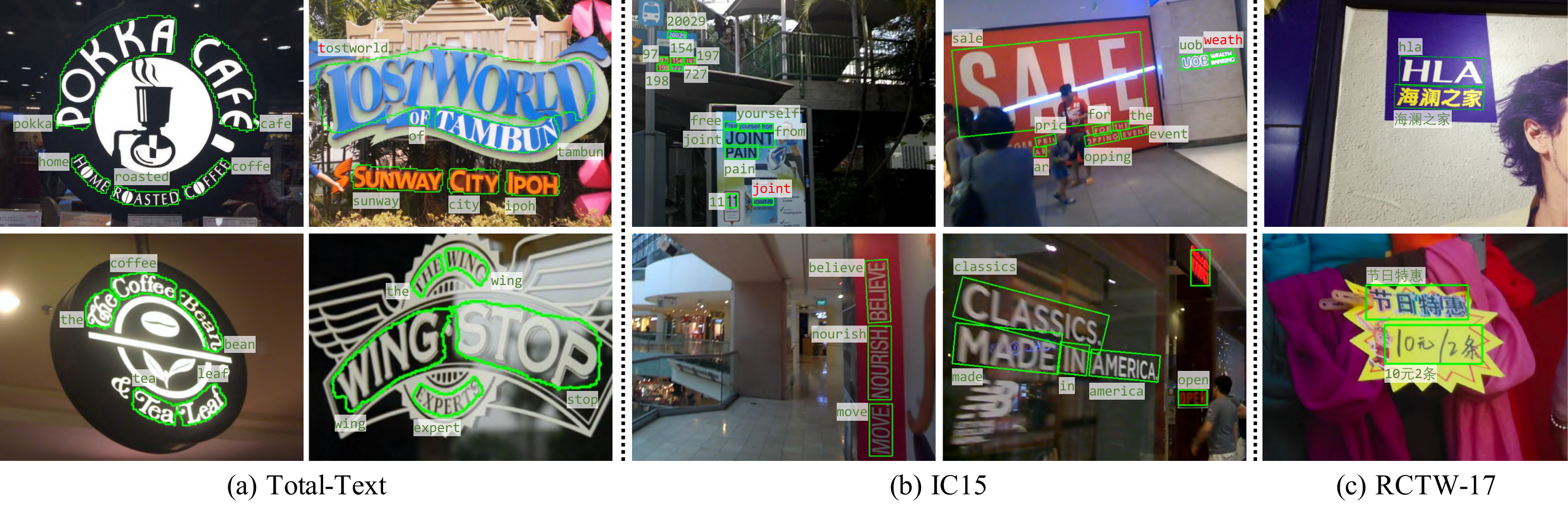}}
		\caption{\textbf{Qualitative text spotting results of PAN++ on Total-Text~\cite{totaltext}, IC15~\cite{karatzas2015icdar}, and RCTW-17~\cite{shi2017icdar2017}.} Our method can accurately detect and recognize arbitrarily-shaped text, even in noisy scenarios.}
		\label{fig:rec_res}
	\end{figure*}

	These results show that in addition to curved text, our PAN++ can also achieve an excellent balance between accuracy and speed when detecting and recognizing straight text. We present some qualitative straight text spotting results in Fig.~\ref{fig:rec_res} (b).
	
	\subsection{\revise{Experiments on Chinese Text Dataset}}
	\subsubsection{\revise{Experiment Setting}}
	\revise{We choose the RCTW-17~\cite{shi2017icdar2017} dataset to test the effectiveness of our method in detecting and recognizing Chinese text,
	which is a commonly used language that is completely different from English.
	Due to the number of character categories is large (up to 3,000), the detection and recognition of Chinese text is challenging.
	RCTW-17 has the same detection metric as English text datasets (\eg, IC15~\cite{karatzas2015icdar}), but it uses the average edit distance (AED)~\cite{shi2017icdar2017} as the recognition metric.
	For a fair comparison, we randomly selected 1,000 images from the RCTW-17 training set as the validation set, and trained all models for 300 epochs in the remaining training set without using external text datasets.
	Other model settings, data augmentations, and training settings are the same as those in Sec. \ref{sec:ct_setting}. 
	In the testing phase, we rescale the shorter side of the input image to 736 pixels. 
	}
	
	\begin{table}[t]
		\centering
		\renewcommand\arraystretch{1}
		\caption{\revise{\textbf{Results on the RCTW-17 dataset}. ``Scale'' denotes the scale of the testing image, where ``\emph{S}:'' means the shorter side is fixed. ``F'' denotes the F-measure. ``AED'' represents average edit distance defined in \cite{shi2017icdar2017}}.}
		\setlength{\tabcolsep}{1.4mm}
		\begin{tabular}{r|c|c|c|c|c}
	\whline
	\multirow{2}{*}{Method} & \multirow{2}{*}{Scale} & \multirow{2}{*}{Backbone} & \multicolumn{3}{c}{RCTW-17} \\
	\cline{4-6}
	& & & F & AED & FPS \\
	\hline
	\hline
	SegLink~\cite{shi2017detecting} + CRNN~\cite{crnn} & 768$\times$1280 & VGG16 & 53.9 & 25.1 & 5.9\\
	\hline
	\multirow{2}{*}{PAN++ (ours)} & \emph{S}: 736 & ResNet18 & 65.1 & 23.4 & \textbf{18.5} \\
	& \emph{S}: 896 & ResNet50 & \textbf{68.0} & \textbf{22.2} & 10.7  \\
	\whline
\end{tabular}
		\label{tab:rctw}
	\end{table}
	
	\subsubsection{\revise{Chinese Text Results}}
	\revise{
	The text detection and recognition performance on the RCTW-17 dataset are reported in Table \ref{tab:rctw}.
	With ResNet18~\cite{he2016deep} as the backbone, our PAN++ obtains
	65.1 text detection F-measure and 23.4 AED, which is 11.2 points (65.1\% \vs 53.9\%) and 1.7 points (23.4\% \vs 25.1\%) better than those of the baseline (SegLink~\cite{shi2017detecting} + CRNN~\cite{crnn}).
	At the same time, our method runs 3 times faster than the baseline (18.5 FPS \vs 5.9 FPS).
	When using a heavier backbone (\ie, ResNet50~\cite{he2016deep}) and a larger input scale, the text detection F-measure and AED of our method further improve to 68.0\% and 22.2\%, 14.1 points (68.0\% \vs 53.9\%) and 2.9 points (22.2\% \vs 25.1\%) better than the baseline, while keeping a faster inference speed (10.7 FPS \emph{vs.} 5.9 FPS).
	Some qualitative Chinese text detection and recognition results are presented in Fig.~\ref{fig:rec_res} (c).
	These results prove that our PAN++ can also works well in Chinese text detection and recognition.
	}

	\subsection{\revise{Comparison with Conference Versions}}
	\revise{
	The key points of PAN++ and its conference versions are summarized as follows:
	\begin{itemize}
	    \item PSENet~\cite{psenet} (\textbf{Text Detector}): Kernel Representation, PSE;
	    \item PAN~\cite{wang2019efficient} (\textbf{Text Detector}): FPEM, PA;
	    \item PAN++ (\textbf{E2E Text Spotter}): Masked RoI, Recognition Head, FPEM (improved), PA (improved);
	\end{itemize}
	As an extension of conference versions, PAN++ evolves from the original text detector to an end-to-end (E2E)
	arbitrarily-shaped text spotter. To this end, we rebuild the overall architecture of PAN++ and carefully integrate a feature extractor (\ie, Masked RoI) and a lightweight recognition head in it.}
	
	\revise{
	Besides that, when inheriting the kernel representation, FPEM, and PA from the conference versions, we make some improvements as follows: 1) We systematically compare our kernel representation with other mainstream representations, and prove that our representation is friendly to real-time applications.
	2) We simplify the FPEM by combining the old FPEM~\cite{wang2019efficient} and FFM~\cite{wang2019efficient} into a single module, which is more effective.
	3) We improve PA and make it aware of background elements by adding a background item to the discrimination loss.
	}
	
	\begin{table}[t]
		\centering
		\renewcommand\arraystretch{1}
		\newcommand{\tabincell}[3]{\begin{tabular}{@{}#1@{}}#2\end{tabular}}
		\caption{\revise{\textbf{Detection performance comparison of PAN++ and its conference versions}. ``\emph{S}:'' means the shorter side is fixed. ``F'' denotes the F-measure. ``impr'' means the improved module in this work. At a similar or faster inference speed, the F-measure of PAN++ is significantly better than previous versions. Note that, the inference speed is tested on one 1080Ti GPU for fair comparisons.}}
		\begin{tabular}{r|c|c|c|c}
	\whline
	\multirow{2}{*}{Method}
	& \multicolumn{2}{c|}{Total-Text (\emph{S}: 640)} & \multicolumn{2}{c}{IC15 (\emph{S}: 736)} \\
	\cline{2-5}
	& F & FPS & F & FPS \\
	\hline
	\hline
	PSENet-1s~\cite{psenet} & 78.1 & 4.5 & 76.9 & 3.8 \\
	\hline
	PAN~\cite{wang2019efficient} & 83.5 & \textbf{39.6} & 80.3 & 26.1 \\
	\hline
	PAN + PA (impr) & 84.0 & 38.2 & 80.7 & 28.0 \\
	\hline
	\multirow{2}{*}{\tabincell{r}{PAN++ (detection only):\\PAN + PA (impr) + FPEM (impr)}} & \multirow{2}{*}{\textbf{84.5}} & \multirow{2}{*}{38.3} & \multirow{2}{*}{\textbf{81.2}} & \multirow{2}{*}{\textbf{28.2}} \\
	& & & &\\
	\whline
\end{tabular}
		\label{tab:cmp}
	\end{table}
	
	\revise{
	With the above-mentioned improvements, PAN++ can simultaneously detect and recognize arbitrarily-shaped text, \emph{which is impossible with models in conference versions}.
	Moreover, PAN++ also achieve better text detection performance than its predecessors. As reported in Table \ref{tab:cmp}, under the same or better inference speed, the F-measure of PAN++ is 84.5\% on the Total-Text dataset, which is 1.0 points better than PAN~\cite{wang2019efficient}, and 6.4 points better than PSENet~\cite{psenet}.
	Similar results are also on the IC15 dataset.
	These results prove that the improvements we made in this work are effective and important for a fast end-to-end text spotter.
	}
	
	\subsection{Ablation Study}
	\subsubsection{Experiment Settings}
	\label{sec:as_setting}
	To analyze PAN++ in depth, we conduct ablation studies on both curved and straight text datasets (\eg, Total-Text and IC15).
	In these experiments, we use ResNet18 as the default backbone network.
	During training, we train text detection models without external text datasets and use the fine-tuning strategy to train text spotting models.
	The shrinking rate $r$ on Total-Text and IC15 is set to 0.7 and 0.5, respectively.
	In the testing phase, the shorter sides of the images in Total-Texts and IC15 are fixed at 640 and 736 pixels, respectively. No lexicon is used to refine the recognition results.
	Other model settings, data augmentations, and training/testing settings follow the settings in Sec. \ref{sec:ct_setting}.
	
	\subsubsection{Can text kernels be used as detection results?}
\begin{figure}[t]
		\begin{center}
			\setlength{\fboxrule}{0pt}
			\fbox{\includegraphics[width=0.48\textwidth]{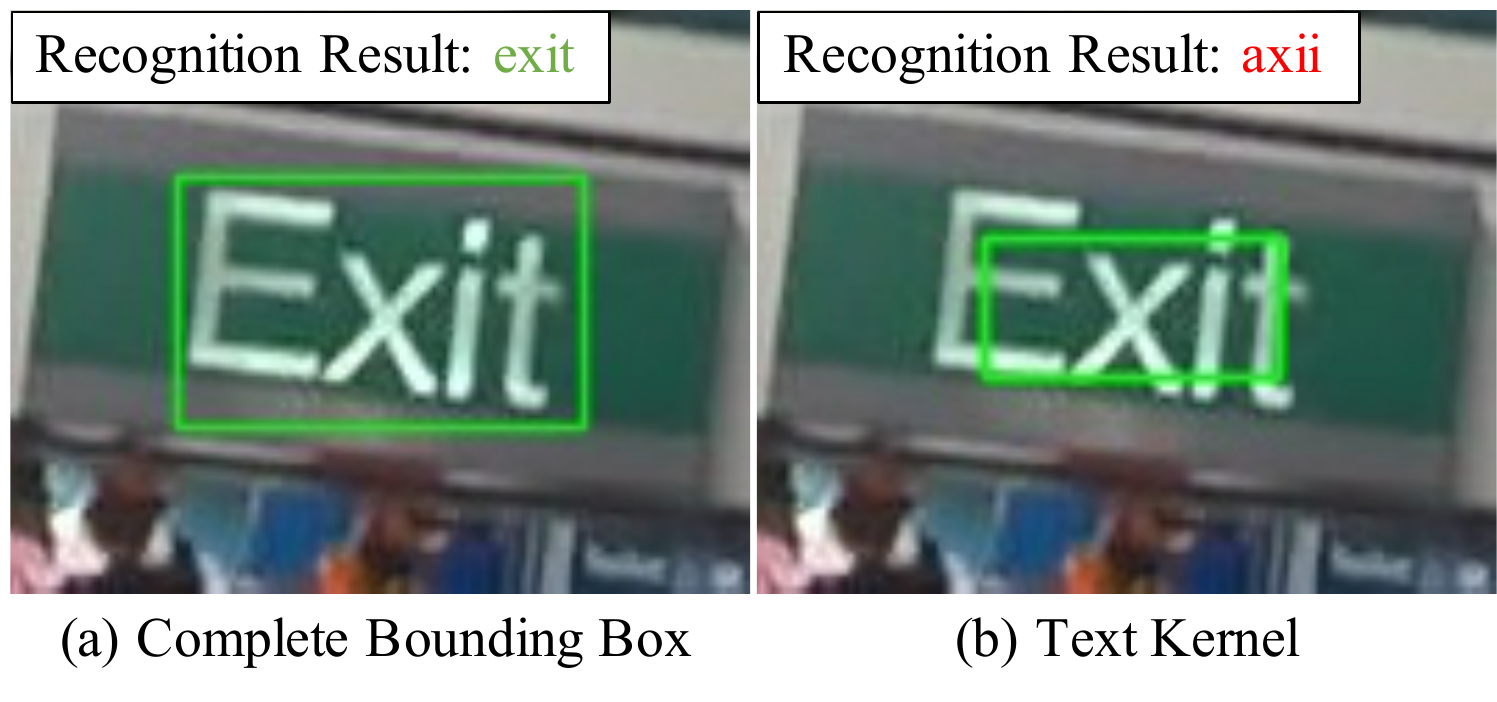}}
			\caption{\textbf{Recognition results of the complete bounding box and the text kernel of the same text line}. CRNN~\cite{crnn} successfully recognizes the text in the complete bounding box but fail to recognize the text in the text kernel.}
			\label{fig:rec_text_kernel}
		\end{center}
	\end{figure}

	As mentioned in Sec.\ref{sec:kernel_rep}, the text kernel is the central region of a text line, which is used to locate and separate adjacent text lines during the segmentation process.
	It cannot cover the complete shape of a text line and cannot be directly used as the detection result.
	As shown in Fig.~\ref{fig:shrink_rate}, the text detection F-measure of the model that only uses text kernel is much lower than the models with the default setting.
	In addition, we use a text recognizer CRNN~\cite{crnn} to recognize the text in the text kernel, and we find that CRNN fails to recognize it as shown in Fig. \ref{fig:rec_text_kernel}. 
	
	These results demostrate that the text kernel cannot be used alone as the detection result, and it needs to be combined with other parts (\ie, text regions and instance vectors) of the kernel representation.
	
	\subsubsection{Influence of the Shrinking Rate}
	\begin{figure}[t]
		\centering
		\setlength{\fboxrule}{0pt}
		\fbox{\includegraphics[width=0.4\textwidth]{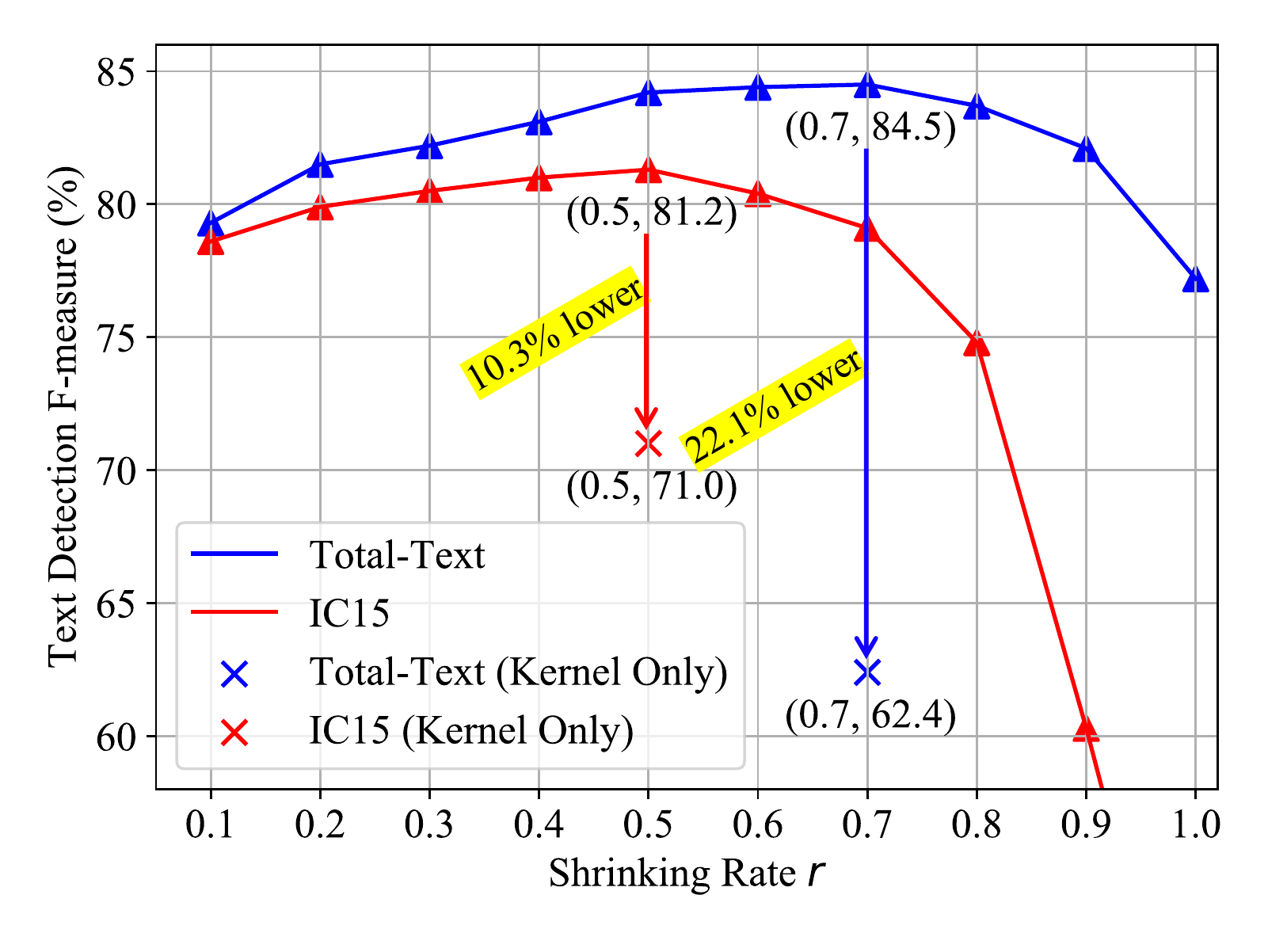}}
		\caption{\textbf{Text detection F-measure of PAN++ under different shrink rates}. When we raise the shrinking rate, the F-measure first increases and then decreases. In addition, the F-measure drops sharply when we only use text kernels as the detection result.
		}
		\label{fig:shrink_rate}
	\end{figure}
	To study the impact of the shrinking rate on the text detection performance, we compare the text detection F-measure of models with different shrinking rates.
	Specifically, we vary the shrinking rate $r$ from 0.1 to 1 and evaluate PAN++ (detection only) on Total-Text and IC15 datasets. 
	As shown in Fig.~\ref{fig:shrink_rate}, the F-measures on both datasets drop when the shrinking rate $r$ is
	too large or too small. 
	When $r$ is too large (\ie, $r\ge0.9$), text kernels may fail to separate the text lines lying close to each other. 
	When $r$ is too small (\ie, $r\le 0.2$), a whole text line may be incorrectly divided into several
	connected components in the testing phase.
	Experimentally, the shrinking rate is set to 0.5 on IC15, and 0.7 on other datasets by default.

	Note that, when the shrinking rate $r$ is equal to 1, the text kernel is the same as the text region, and the detection module will degenerate to a binary segmentation network without PA 
	In this case, the text detection F-measure drops sharply on Total-Text and IC15, because the model loses the capability to distinguish adjacent text lines as shown in Fig.~\ref{fig:rep} (b).
	
	\subsubsection{Influence of the Stack Number of FPEMs}
	We study the effect of the stack number of FPEMs on text spotting performance, by varying the stack number $N_{stk}$ from $0$ to $4$.
	Note that, when the stack number $N_{stk}$ is equal to 0,  there is no FPEM and the final feature map $F_f$ is generated by upsampling and concatenating the feature pyramid $F_r$ (see Fig.~\ref{fig:arch} (c)).
	From Table~\ref{tab:abs_fpem}, we find that the text spotting F-measures on both datasets keep rising with the growth of $N_{stk}$ and begins to level off when the stack number $N_{stk}$ is greater than 2.
	The stack number $N_{stk}$ is a trade-off between accuracy and speed.
	Although FPEM has a low computational cost, the model will slow down when the stack number $N_{stk}$ is large. 
	Specifically, an additional FPEM takes up to 2ms per image.
	To keep a good balance between performance and speed, $N_{stk}$ is set to 2 by default.
	
	\begin{table}[t]
		\centering
		\renewcommand\arraystretch{1}
		\newcommand{\tabincell}[3]{\begin{tabular}{@{}#1@{}}#2\end{tabular}}
		\caption{\textbf{End-to-end text spotting F-measures of models with different number of stacked FPEMs}. ``Param.'' represents the number of parameters of the entire model. \revise{``F'' denotes the F-measure. ``Time'' refers to the time cost of FPEMs per image.}}
		\begin{tabular}{c|c|c|c|c|c}
	\whline
	\multirow{2}{*}{$N_{stk}$} & 
	\multirow{2}{*}{Param. (M)}
	& \multicolumn{2}{c|}{Total-Text} & \multicolumn{2}{c}{IC15} \\
	\cline{3-6}
	& & F & Time (ms) & F & Time (ms) \\
	\hline
	\hline
	0 & \textbf{13.1} & 64.5 & \textbf{0.6} & 63.7 & \textbf{0.7}\\
	\hline
	1 & 13.2 & 66.0 & 2.2 & 64.2 & 2.8 \\
	\hline
	2 & 13.3 & 66.4 & 3.8 & 64.7 & 4.4 \\
	\hline
	3 & 13.4 & 66.5 & 5.1 & 64.8 & 6.6\\
	\hline
	4 & 13.5 & \textbf{66.6} & 6.5 & \textbf{64.9} & 8.7 \\
	\whline
\end{tabular}
		\label{tab:abs_fpem}
	\end{table}
	
	\subsubsection{Effectiveness of FPEM}
	To verify the effectiveness of FPEM, we first compare the performance of models with and without this module.
	As reported in Table~\ref{tab:abs_fpem}, the model with 2-stacked FPEMs ($N_{stk} = 2$) obtains the F-measure of 66.4\% on the Total-Text dataset, which is 1.9 points higher that of the model without FPEM ($N_{stk} = 0$).
	
	Second, we compare our feature enhancement network (2-stacked FPEMs) with other widely used methods, including FPN~\cite{lin2017feature}, PAFPN~\cite{liu2018path}, and NAS-FPN~\cite{ghiasi2019fpn}.
	For fair comparisons, we directly replace the feature enhancement network with other methods and evaluate them under the same experiment setting mentioned in Sec. \ref{sec:as_setting}.
	\revise{
	In Table \ref{tab:abs_pspnet}, we find that our method achieves comparable or even better text spotting performance than counterparts, while our time cost and parameter number are much lower.
	Concretely, our 2-stacked FPEMs yields the end-to-end text spotting F-measure of 66.4\%  on Total-Text, which is better than FPN~\cite{lin2017feature} (65.9\%) and close to NAS-FPN~\cite{ghiasi2019fpn} (66.9\%). However, the time cost of our method is only 1/3 of FPN and 1/10 of NAS-FPN (3.8 ms \vs 12.1 ms and 41.2 ms), and our parameters is 1/2 of FPN and 1/4 of NAS-FPN (13.3 M \vs 20.3 M and 46.9 M).
	}
	
    In conclusion, the proposed FPEM can obtain a good balance between accuracy and inference speed/parameter number, benefiting from the separable convolution and stackable design.
	It is a good choice to enhance the feature from a lightweight backbone, at least for text detection and end-to-end text spotting tasks.
	
	\begin{table}[t]
		\centering
		\renewcommand\arraystretch{1}
		\caption{\textbf{Comparison of different feature enhancement networks.} ``Param.'' represents the number of parameters of the entire model. \revise{``F'' denotes the F-measure. ``Time'' refers to the time cost of the feature enhancement network per image.}}
		\setlength{\tabcolsep}{0.9mm}
		
\begin{tabular}{r|c|c|c|c|c}
	\whline
	\multirow{2}{*}{Method} & \multirow{2}{*}{Param. (M)} & \multicolumn{2}{c|}{Total-Text} & \multicolumn{2}{c}{IC15} \\
	\cline{3-6}
	& & F & Time (ms) & F & Time (ms) \\
	\hline
	\hline
	2-Stacked FPEMs (ours) & \textbf{13.3} & 66.4 & \textbf{3.8} & 64.7 & \textbf{4.4} \\
	\hline
	FPN~\cite{lin2017feature} & 20.3 & 65.9 & 12.1 & 64.5 & 15.6 \\
	\hline
	PAFPN~\cite{liu2018path} & 23.8 & 66.2 & 13.4 & 64.7 & 17.7 \\
	\hline
	NAS-FPN~\cite{ghiasi2019fpn} & 46.9 & \textbf{66.9} & 41.2 & \textbf{65.1} & 54.3 \\
	\whline
\end{tabular}
		\label{tab:abs_pspnet}
	\end{table}
	
	\subsubsection{Effectiveness of PA}
	\begin{figure}[t]
		\centering
		\setlength{\fboxrule}{0pt}
		\fbox{\includegraphics[width=0.48\textwidth]{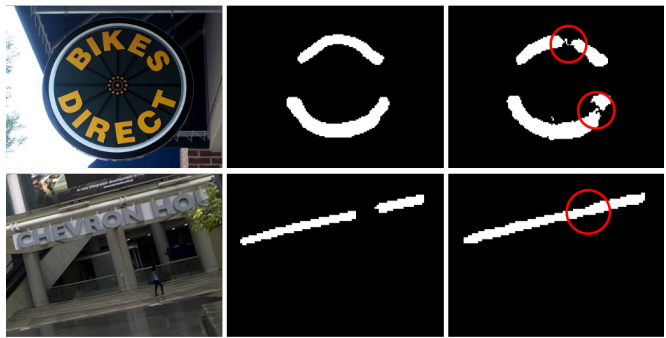}}
		\caption{\textbf{Text kernel results of models with and without PA}. The model with PA tends to predict better text kernels, reducing the incorrect cases shown in red circles.}
		\label{fig:pa}
	\end{figure}
	We study the effectiveness of PA by removing it from the pipeline. In detail, we set $\beta$ in Eqn.~\eqref{eqn:loss_det} to 0 during training and merge all neighbor text pixels in step 2 of post-processing.
	Comparing the model with PA (see Table~\ref{tab:abs_many}~\#1), the F-measure of the model without PA (see Table~\ref{tab:abs_many}~\#2) drops over 1.0 points, which indicates the effectiveness of PA.
	
	In Fig. \ref{fig:pa}, we also present some text kernel results predicted by models with and without PA. We find that the model with PA tends to predict higher-quality text kernels, which can effectively reduce the incorrect cases including 1) dividing a text line into several parts and 2) connecting two adjacent text lines 
	
	From these results, we claim that PA has two benefits: 1) As an auxiliary loss (see Eqns.~\eqref{eqn:loss_agg} and~\eqref{eqn:loss_dis}), it can improve the segmentation result of the model and make it more compact and accurate; 2) As a post-processing step (see Fig.~\ref{fig:dis_infer}), it can deal with the conflict pixel problem by aggregating conflict pixels to the correct text kernel.
	
	\begin{table}[t]
		\centering
		\renewcommand\arraystretch{1}
		\caption{\textbf{Comparison of the models with and without PA, and with different backbone networks.} \revise{``FPS'' denotes the inference speed of the entire model.}}
		\begin{tabular}{c|c|c|c|c|c|c}
	\whline
	\multirow{2}{*}{\#} &
	\multirow{2}{*}{Backbone} &
	\multirow{2}{*}{w/ PA} & 
	\multicolumn{2}{c|}{Total-Text} & \multicolumn{2}{c}{IC15} \\
	\cline{4-7}
	& & & F-measure & FPS & F-measure & FPS \\
	\hline
	\hline
	1 & ResNet18 & \checkmark & 66.4 & 24.1 & 64.7 & 24.5 \\
	2 & ResNet18 & $\times$ & 64.9 & \textbf{24.9} & 62.9 & \textbf{25.0} \\
	\hline
	3 & ResNet50 & \checkmark & 67.2 & 20.2 & 65.0 & 18.1 \\
	\hline
	4 & VGG16 & \checkmark & \textbf{67.8} & 9.2 & \textbf{66.2} & 8.7 \\
	\whline
\end{tabular}

		\label{tab:abs_many}
	\end{table}
	
	\subsubsection{Effectiveness of Masked RoI}
	As shown in Table~\ref{tab:rec} \#1 and \#2, we show the effectiveness of Masked RoI. 
	Higher text spotting F-measure is achieved in the presence of Masked RoI. 
	The model with the proposed Masked RoI outperforms the one with the naive RoI extractor (removing the masking operation in Masked RoI) by at least 1.1\%. 
	The underlying reason is that without Masked RoI, the model may fail to recognize text lines in a noisy background.
	
	\subsubsection{Effectiveness of Recognition Head}
	To test the effectiveness of our recognition head, we compare it with some state-of-the-art methods, including SAR~\cite{li2019show} and Transformer~\cite{vaswani2017attention}. For fair comparisons, we replace our recognition head with other methods and evaluate them under the same experiment setting mentioned in Sec. \ref{sec:as_setting}.
	\revise{As shown in Table \ref{tab:rec}, our recognition head outperform SAR~\cite{li2019show} in terms of accuracy and inference speed. Especially in terms of speed, the time cost of our method is only 1/2 of SAR's (2.3ms \emph{vs} 5.0ms), which is important when encountering the image with dense text. Compared with Transformer~\cite{vaswani2017attention}, our method yields a comparable F-measure (66.4\% \vs 67.0\%) at a speed of 2.3ms per text line, which is 10 times faster.}
	
	These results demonstrate that our recognition head keeps a good balance between accuracy and inference speed. It may benefit from the designs as follows:
	1) Unlike SAR and Transformer, our method has no encoder, skipping the time cost of the encoder;
	2) We adopt the multi-head attention layer, whose computational cost is lower than that of SAR; 
	3) Our decoder only contains two LSTM layers and one attention layer.
	
	\subsubsection{Influence of the Backbone Network}
	To analyze the capability of the proposed PAN++, we replace the lightweight backbone to heavier backbone (\ie, ResNet50 and VGG16). As shown in Table~\ref{tab:abs_many}~\#1, \#3, and \#4, heavier backbones can bring over 1 points improvement on both Total-Text and IC15 datasets.
	However, the reduction of FPS brought by the heavy backbone is also apparent.
	
	\begin{table}[t]
		\centering
		\renewcommand\arraystretch{1}
		\caption{\revise{\textbf{Comparison between the end-to-end text spotting framework and the separate framework.} The end-to-end framework is superior to the separate framework in terms of text detection (Det) F-measure, end-to-end (E2E) text spotting F-measure, and inference speed.}}
		\begin{tabular}{r|c|c|c}
	\whline
	\multirow{2}{*}{Framework} & 
	\multicolumn{3}{c}{Total-Text} \\
	\cline{2-4}
	& F-measure (Det) & F-measure (E2E) & FPS \\
	\hline
	\hline
	End-to-End (ours) & 86.0 & 66.4 & 24.1 \\
	\hline
	Separate & 84.9 & 63.4 & 16.0 \\
	\whline
\end{tabular}

		\label{tab:e2e}
	\end{table}
	
	\subsubsection{\revise{End-to-End Framework \emph{vs.} Separate Framework}}
	\revise{
	To prove the superiority of the end-to-end text spotting framework, we compare the performance of the end-to-end text spotting model with a separate model comprised of a independent detector and recognizer.
	For fair comparisons, we separate the end-to-end PAN++ into a text detector and a text recognizer, and each of them has its own feature encoding network (\ie, ResNet18 + 2-stacked FPEMs).
	}
	
	\revise{
	As reported in Table \ref{tab:e2e}, under the same experiment settings, the end-to-end model achieve 86.0\% text detection (Det) F-measure and 66.4\% end-to-end (E2E) text spotting F-measure, which are 1.1 and 3.0 points better than the separate model. Meanwhile, the inference speed of the end-to-end model has obvious advantages over the separate model.
	}
	
	\revise{
	These results indicate that the end-to-end text spotting framework is much better than the separate framework in terms of accuracy and inference speed.
	The potential reasons are as follows:
	1) In the end-to-end framework, the detection module and the recognition module are jointly trained, so the backbone features can be optimized by the detection loss and the recognition loss simultaneously. As a result, the feature representation capacity in the end-to-end framework is stronger than the separate framework, which effectively reduces false-positive results.
	2) Due to the shared backbone, The end-to-end framework skips the time cost of image feature encoding in text recognition, so its speed is much faster than that of the separate framework.
	}

	\begin{table}[t]
		\centering
		\renewcommand\arraystretch{1}
		\caption{\textbf{Comparison of our recognition head (with and without Masked RoI), SAR~\cite{li2019show}, and Transformer~\cite{vaswani2017attention}}. \revise{``F'' denotes the F-measure. ``Time'' refers to the time cost of the recognizer per text line.}}
		\setlength{\tabcolsep}{1.1mm}
		\begin{tabular}{c|c|c|c|c|c|c}
	\whline
	\multirow{2}{*}{\#} &
	\multirow{2}{*}{Rec. Head} &
	\multirow{2}{*}{w/ Masked RoI} & 
	\multicolumn{2}{c|}{Total-Text} & \multicolumn{2}{c}{IC15} \\
	\cline{4-7}
	& & & F & Time (ms) & F & Time (ms) \\
	\hline
	\hline
	1 & Ours & \checkmark & 66.4 & \textbf{2.3} & 64.7 & \textbf{2.6}\\
	2 & Ours & $\times$ & 65.1 & \textbf{2.3} & 63.6 & \textbf{2.6} \\
	\hline
	3 & SAR~\cite{li2019show} & \checkmark & 66.1 & 5.0 & 64.2 & 5.2 \\
	\hline
	4 & Transformer~\cite{vaswani2017attention} & \checkmark & \textbf{67.0} & 19.5 & \textbf{65.2} & 20.1 \\
	\whline
\end{tabular}

		\label{tab:rec}
	\end{table}
	
	\begin{figure*}[t]
		\centering
		\setlength{\fboxrule}{0pt}
		\fbox{\includegraphics[width=0.9\textwidth]{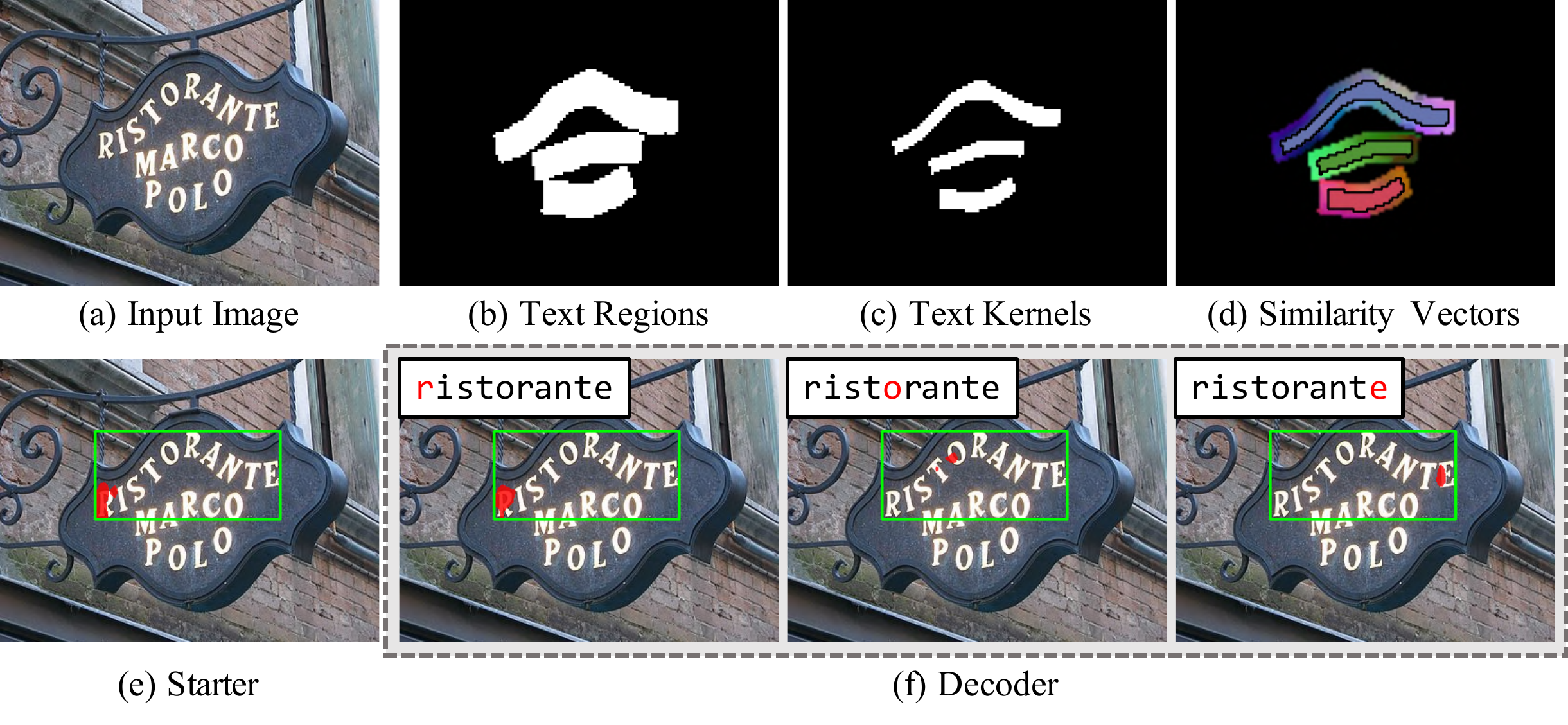}}
		\caption{Visualization of intermediate results produced by the detection head and the recognition head.}
		\label{fig:det_vis}
	\end{figure*}
	
	\subsection{Network Output Visualization}
	Some intermediate results of the model are shown in Fig.~\ref{fig:det_vis}, which includes the results of text regions, text kernels, instance vectors, and attention masks.
	We find that the text regions (see Fig.~\ref{fig:det_vis}~(b)) keep the complete shapes of text lines, and text kernels (see Fig.~\ref{fig:det_vis}~(c)) can distinguish different text lines clearly. In Fig.~\ref{fig:det_vis}~(d), we reduce the dimension of
	the instance vector to 3 through PCA and project it to the RGB field. We find that pixels and text kernels belonging to a text line tend to have similar colors.
	
	Fig.~\ref{fig:det_vis} (e) shows the attention mask generated by the recognition head's starter, which accurately locates the start of the string.
	Fig.~\ref{fig:det_vis} (f) are attention masks of the recognition head's decoder at each step, 
	which provides a great tool to analyze the performance of the recognition head. We find that the attention masks of the decoder can focus on the correct position even when recognizing the irregular text line.

	\subsection{Robustness Analysis}
	To further demonstrate the robustness of the proposed PAN++, we evaluate the model without recognition head trained on one dataset and test on other datasets.
    Because this experiment aims to test the generalization of our method, we do not use external datasets during training, and we also remove the recognition head that needs to be trained on the large-scale dataset. 
    In the testing phase, we set the shorter sides of test images in Total-Text, CTW1500, IC15, and MSRA-TD500 datasets to 640, 640, 736, and 736, respectively.
	
	Based on the annotation level, we divide the datasets into two groups, which are word level and text line level datasets.
	SynthText, ICDAR 2015, and Total-Text are annotated at the word level, and CTW1500 and MSRA-TD500 are annotated at the text line level.
	
    The cross-dataset results of PAN++ are shown in Table~\ref{tab:cross}.
    Notably, the model trained on SynthText (a synthetic dataset) has fairish performances on IC15 and Total-Text, which indicates that even without any manually annotated data, our method can satisfy the scene text detection with low precision requirements.
    The model trained on manually annotated dataset has over 62 text detection F-measure in the cross-dataset evaluation, which is competitive.
	Furthermore, in the cross-dataset evaluation at the text line level, the F-measure of all models exceeds 78, even though these models are trained on curved/straight text datasets and tested on another dataset. These results demonstrate that the detection part of PAN++ can be well generalized to unseen datasets.

	\begin{table}[t]
		\centering
		\renewcommand\arraystretch{1}
		\caption{\textbf{Cross-dataset results of PAN on word-level and line-level datasets}. 
		}
		\setlength{\tabcolsep}{0.9mm}
		\newcommand{\tabincell}[2]{\begin{tabular}{@{}#1@{}}#2\end{tabular}}
\begin{tabular}{r|c|c|c|c}
	\whline
	Annotation & Train Set$\rightarrow$Test Set & Precision & Recall & F-measure\\
	\hline
	\hline
	\multirow{4}{*}{\tabincell{c}{Word Level}} & SynthText$\rightarrow$IC15 & 68.2 & 44.6 & 53.9 \\
	& SynthText$\rightarrow$Total-Text & 71.5 & 40.2 & 51.5 \\
	& IC15$\rightarrow$Total-Text & 71.1 & 56.4 & 62.9\\
	& Total-Text$\rightarrow$IC15 & 78.7 & 68.6 & 73.3\\
	\hline
	\multirow{2}{*}{\tabincell{c}{Text Line Level}} & CTW1500$\rightarrow$MSRA-TD500 & 80.3 & 79.0 & 79.7\\
	& MSRA-TD500$\rightarrow$CTW1500 & 84.1 & 73.0 & 78.2\\
	\whline
\end{tabular}
		\label{tab:cross}
	\end{table}
	
	\subsection{Speed Analysis}
    Table~\ref{tab:speed} present the time cost of all components of PAN++. We find that the time cost of text recognition accounts for half of the total time cost. Therefore, an obvious way to increase speed in practical applications is to run the recognition component in parallel through a basic producer-consumer model, which can reduce the time cost to the original 1/2.
	We conduct the speed analysis on Total-Text. We evaluate all testing images and calculate the average speed. These results are tested with 1 batch size on one V100 GPU and one 2.20GHz CPU in a single thread.
	
	\begin{table}[t]
		\centering
		\renewcommand\arraystretch{1}
		\caption{\textbf{Time cost of PAN++ on the Total-Text dataset}. The total time cost of backbone network, 2-stacked FPEMs, text detection (with PA), Text Recognition.}
		\setlength{\tabcolsep}{1.3mm}
		\begin{tabular}{r|c|c|c|c|c}
	\whline
	\multirow{2}{*}{Method} & \multicolumn{4}{c|}{Time consumption~(ms)} & \multirow{2}{*}{FPS} \\
	\cline{2-5}
	& Backbone & FPEMs  & Detection (w/ PA) & Recognition \\
	\hline
	PAN++ 512 & 5.2 & 3.0 & 1.4 + PA: 3.5 & 21.2 & 29.2 \\
	\hline
	PAN++ 640 & 7.1 & 3.8 & 2.1 + PA: 5.8 & 22.7 & 24.1 \\
	\hline
	PAN++ 736 & 9.0 & 4.4 & 2.8 + PA: 7.6 & 23.6 & 21.1 \\
	
	\whline
\end{tabular}

		\label{tab:speed}
	\end{table}
	
	\subsection{Failure Cases and Discussion}
	\begin{figure}[t]
		\centering
		\setlength{\fboxrule}{0pt}
		\fbox{\includegraphics[width=0.48\textwidth]{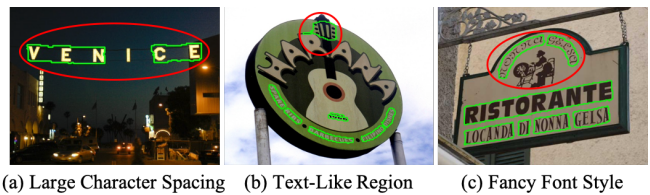}}
		\caption{Failure Samples.}
		\label{fig:failure}
	\end{figure}
	
	As demonstrated in previous experiments, the proposed PAN++ works well in most cases of arbitrarily-shaped text detection and recognition.
	It still fails for some difficult images, such as detecting large character spacing (see Fig.~\ref{fig:failure}~(a)) or text-like areas (see Fig.~\ref{fig:failure}~(b)) and recognizing texts in fancy font styles (see Fig.~\ref{fig:failure}~(c). 
	Large character spacing is a problem widely existing in other state-of-the-art methods, which needs linguistic features to solve it.
	For text-like areas and text in fancy font styles, the main reason for them is the lack of training samples, and we believe that these problems will alleviate when we have sufficient training samples of them.

	\section{Conclusion}
	\revise{
	In this work, we %
	have considerably extended the kernel representation for text spotting, which first appeared in 
	our conference versions. We have also thoroughly discussed  
	its feasibility in real-time arbitrarily-shaped text recognition.
	}
    In particular, 
    To %
    further strengthen this 
    representation, here we have developed an efficient framework for end-to-end arbitrarily-shaped text spotting, where we %
    have managed to speed up
    all parts in the text spotting procedure %
    by carefully designing 
    a series of lightweight modules, including 1) a feature enhancement network with stacked FPEMs, 2) a detection head with PA, and 3) a recognition head with Masked RoI.
    
    \revise{
    We 
    have
    verified the performance of PAN++ on both text detection and end-to-end text spotting tasks.
    The experiment results demonstrate that our method %
    offers
    advantages in both accuracy and inference speed compared with %
    a few
    previous state-of-the-art methods. We hope that our PAN++ can serve as a cornerstone for real-world text understanding applications.
    }
	
	\section*{Acknowledgments}
	This work was supported by the Natural Science Foundation of China under Grant 61672273 and Grant 61832008, the Science Foundation for Distinguished Young Scholars of Jiangsu under Grant BK20160021. This work was in part 
	supported by Alibaba Group through Alibaba Innovative Research (AIR) Program.
	
	\ifCLASSOPTIONcaptionsoff
	\newpage
	\fi
	
	{\small
		\bibliographystyle{ieeetr}
		\bibliography{egbib}

\begin{thebibliography}{10}

\bibitem{simonyan2014very}
K.~Simonyan and A.~Zisserman, ``Very deep convolutional networks for
  large-scale image recognition,'' {\em arXiv preprint arXiv:1409.1556}, 2014.

\bibitem{he2016identity}
K.~He, X.~Zhang, S.~Ren, and J.~Sun, ``Identity mappings in deep residual
  networks,'' in {\em Proc. Eur. Conf. Comp. Vis.}, 2016.

\bibitem{zhou2017east}
X.~Zhou, C.~Yao, H.~Wen, Y.~Wang, S.~Zhou, W.~He, and J.~Liang, ``{EAST}: an
  efficient and accurate scene text detector,'' in {\em Proc. IEEE Conf. Comp.
  Vis. Patt. Recogn.}, 2017.

\bibitem{psenet}
W.~Wang, E.~Xie, X.~Li, W.~Hou, T.~Lu, G.~Yu, and S.~Shao, ``Shape robust text
  detection with progressive scale expansion network,'' in {\em Proc. IEEE
  Conf. Comp. Vis. Patt. Recogn.}, 2019.

\bibitem{crnn}
B.~Shi, X.~Bai, and C.~Yao, ``An end-to-end trainable neural network for
  image-based sequence recognition and its application to scene text
  recognition,'' {\em {IEEE} Trans. Pattern Anal. Mach. Intell.}, vol.~39,
  no.~11, pp.~2298--2304, 2016.

\bibitem{shi2018aster}
B.~Shi, M.~Yang, X.~Wang, P.~Lyu, C.~Yao, and X.~Bai, ``Aster: An attentional
  scene text recognizer with flexible rectification,'' {\em {IEEE} Trans.
  Pattern Anal. Mach. Intell.}, vol.~41, no.~9, pp.~2035--2048, 2018.

\bibitem{liu2018fots}
X.~Liu, D.~Liang, S.~Yan, D.~Chen, Y.~Qiao, and J.~Yan, ``{FOTS}: Fast oriented
  text spotting with a unified network,'' in {\em Proc. IEEE Conf. Comp. Vis.
  Patt. Recogn.}, 2018.

\bibitem{masktextspotter}
P.~Lyu, M.~Liao, C.~Yao, W.~Wu, and X.~Bai, ``Mask textspotter: An end-to-end
  trainable neural network for spotting text with arbitrary shapes,'' in {\em
  Proc. Eur. Conf. Comp. Vis.}, 2018.

\bibitem{textboxes++}
M.~Liao, B.~Shi, and X.~Bai, ``Textboxes++: A single-shot oriented scene text
  detector,'' {\em {IEEE} Trans. Image Process.}, vol.~27, no.~8,
  pp.~3676--3690, 2018.

\bibitem{FCN}
J.~Long, E.~Shelhamer, and T.~Darrell, ``Fully convolutional networks for
  semantic segmentation,'' in {\em Proc. IEEE Conf. Comp. Vis. Patt. Recogn.},
  2015.

\bibitem{spcnet}
E.~Xie, Y.~Zang, S.~Shao, G.~Yu, C.~Yao, and G.~Li, ``Scene text detection with
  supervised pyramid context network,'' in {\em Proc. {AAAI} Conf. Artificial
  Intell.}, 2019.

\bibitem{tian2016detecting}
Z.~Tian, W.~Huang, T.~He, P.~He, and Y.~Qiao, ``Detecting text in natural image
  with connectionist text proposal network,'' in {\em Proc. Eur. Conf. Comp.
  Vis.}, 2016.

\bibitem{li2019show}
H.~Li, P.~Wang, C.~Shen, and G.~Zhang, ``Show, attend and read: A simple and
  strong baseline for irregular text recognition,'' in {\em Proc. {AAAI} Conf.
  Artificial Intell.}, 2019.

\bibitem{li2017towards}
H.~Li, P.~Wang, and C.~Shen, ``Towards end-to-end text spotting with
  convolutional recurrent neural networks,'' in {\em Proc. IEEE Int. Conf.
  Comp. Vis.}, 2017.

\bibitem{he2018end}
T.~He, Z.~Tian, W.~Huang, C.~Shen, Y.~Qiao, and C.~Sun, ``An end-to-end
  textspotter with explicit alignment and attention,'' in {\em Proc. IEEE Conf.
  Comp. Vis. Patt. Recogn.}, 2018.

\bibitem{qin2019towards}
S.~Qin, A.~Bissacco, M.~Raptis, Y.~Fujii, and Y.~Xiao, ``Towards unconstrained
  end-to-end text spotting,'' in {\em Proc. IEEE Int. Conf. Comp. Vis.}, 2019.

\bibitem{wang2019efficient}
W.~Wang, E.~Xie, X.~Song, Y.~Zang, W.~Wang, T.~Lu, G.~Yu, and C.~Shen,
  ``Efficient and accurate arbitrary-shaped text detection with pixel
  aggregation network,'' in {\em Proc. IEEE Int. Conf. Comp. Vis.}, 2019.

\bibitem{maskrcnn}
K.~He, G.~Gkioxari, P.~Doll{\'a}r, and R.~Girshick, ``Mask {R-CNN},'' in {\em
  Proc. IEEE Int. Conf. Comp. Vis.}, 2017.

\bibitem{he2016deep}
K.~He, X.~Zhang, S.~Ren, and J.~Sun, ``Deep residual learning for image
  recognition,'' in {\em Proc. IEEE Conf. Comp. Vis. Patt. Recogn.}, 2016.

\bibitem{hochreiter1997long}
S.~Hochreiter and J.~Schmidhuber, ``Long short-term memory,'' {\em Neural
  computation}, vol.~9, no.~8, pp.~1735--1780, 1997.

\bibitem{vaswani2017attention}
A.~Vaswani, N.~Shazeer, N.~Parmar, J.~Uszkoreit, L.~Jones, A.~N. Gomez,
  {\L}.~Kaiser, and I.~Polosukhin, ``Attention is all you need,'' in {\em Proc.
  Advances in Neural Inf. Process. Syst.}, 2017.

\bibitem{liu2020abcnet}
Y.~Liu, H.~Chen, C.~Shen, T.~He, L.~Jin, and L.~Wang, ``Abcnet: Real-time scene
  text spotting with adaptive bezier-curve network,'' in {\em Proc. IEEE Conf.
  Comp. Vis. Patt. Recogn.}, 2020.

\bibitem{totaltext}
C.~K. Ch'ng and C.~S. Chan, ``Total-text: A comprehensive dataset for scene
  text detection and recognition,'' in {\em Proc. Int. Conf. Document Analysis
  Recogn.}, 2017.

\bibitem{Liu2017Detecting}
L.~Yuliang, J.~Lianwen, Z.~Shuaitao, and Z.~Sheng, ``Detecting curve text in
  the wild: New dataset and new solution,'' {\em arXiv preprint
  arXiv:1712.02170}, 2017.

\bibitem{karatzas2015icdar}
D.~Karatzas, L.~Gomez-Bigorda, A.~Nicolaou, S.~Ghosh, A.~Bagdanov, M.~Iwamura,
  J.~Matas, L.~Neumann, V.~R. Chandrasekhar, S.~Lu, {\em et~al.}, ``Icdar 2015
  competition on robust reading,'' in {\em Proc. Int. Conf. Document Analysis
  Recogn.}, 2015.

\bibitem{msra}
C.~Yao, X.~Bai, W.~Liu, Y.~Ma, and Z.~Tu, ``Detecting texts of arbitrary
  orientations in natural images,'' in {\em Proc. IEEE Conf. Comp. Vis. Patt.
  Recogn.}, 2012.

\bibitem{ren2015faster}
S.~Ren, K.~He, R.~Girshick, and J.~Sun, ``{Faster R-CNN}: Towards real-time
  object detection with region proposal networks,'' in {\em Proc. Advances in
  Neural Inf. Process. Syst.}, 2015.

\bibitem{busta2017deep}
M.~Busta, L.~Neumann, and J.~Matas, ``Deep textspotter: An end-to-end trainable
  scene text localization and recognition framework,'' in {\em Proc. IEEE Int.
  Conf. Comp. Vis.}, 2017.

\bibitem{nguyenvan2019pooling}
D.~NguyenVan, S.~Lu, S.~Tian, N.~Ouarti, and M.~Mokhtari, ``A pooling based
  scene text proposal technique for scene text reading in the wild,'' {\em
  Pattern Recogn.}, vol.~87, pp.~118--129, 2019.

\bibitem{liao2019mask}
M.~Liao, P.~Lyu, M.~He, C.~Yao, W.~Wu, and X.~Bai, ``Mask textspotter: An
  end-to-end trainable neural network for spotting text with arbitrary
  shapes,'' {\em {IEEE} Trans. Pattern Anal. Mach. Intell.}, 2019.

\bibitem{xing2019charnet}
L.~Xing, Z.~Tian, W.~Huang, and M.~R. Scott, ``Convolutional character
  networks,'' in {\em Proc. IEEE Int. Conf. Comp. Vis.}, 2019.

\bibitem{feng2019textdragon}
W.~Feng, W.~He, F.~Yin, X.-Y. Zhang, and C.-L. Liu, ``Textdragon: An end-to-end
  framework for arbitrary shaped text spotting,'' in {\em Proc. IEEE Int. Conf.
  Comp. Vis.}, 2019.

\bibitem{liao2017textboxes}
M.~Liao, B.~Shi, X.~Bai, X.~Wang, and W.~Liu, ``Textboxes: A fast text detector
  with a single deep neural network,'' in {\em Proc. {AAAI} Conf. Artificial
  Intell.}, 2017.

\bibitem{shi2017detecting}
B.~Shi, X.~Bai, and S.~Belongie, ``Detecting oriented text in natural images by
  linking segments,'' in {\em Proc. IEEE Conf. Comp. Vis. Patt. Recogn.}, 2017.

\bibitem{tian2015text}
S.~Tian, Y.~Pan, C.~Huang, S.~Lu, K.~Yu, and C.~Lim~Tan, ``Text flow: A unified
  text detection system in natural scene images,'' in {\em Proc. IEEE Int.
  Conf. Comp. Vis.}, 2015.

\bibitem{PixelLink}
D.~Deng, H.~Liu, X.~Li, and D.~Cai, ``{PixelLink}: Detecting scene text via
  instance segmentation,'' in {\em Proc. {AAAI} Conf. Artificial Intell.},
  2018.

\bibitem{textsnake}
S.~Long, J.~Ruan, W.~Zhang, X.~He, W.~Wu, and C.~Yao, ``{TextSnake}: A flexible
  representation for detecting text of arbitrary shapes,'' {\em Proc. Eur.
  Conf. Comp. Vis.}, 2018.

\bibitem{xue2019msr}
C.~Xue, S.~Lu, and W.~Zhang, ``{MSR}: Multi-scale shape regression for scene
  text detection,'' in {\em Proc. Int. Joint Conf. Artificial Intell.}, 2019.

\bibitem{graves2006connectionist}
A.~Graves, S.~Fern{\'a}ndez, F.~Gomez, and J.~Schmidhuber, ``Connectionist
  temporal classification: labelling unsegmented sequence data with recurrent
  neural networks,'' in {\em Proc. Int. Conf. Mach. Learn.}, 2006.

\bibitem{bissacco2013photoocr}
A.~Bissacco, M.~Cummins, Y.~Netzer, and H.~Neven, ``Photoocr: Reading text in
  uncontrolled conditions,'' in {\em Proc. IEEE Int. Conf. Comp. Vis.}, 2013.

\bibitem{jaderberg2014deep}
M.~Jaderberg, A.~Vedaldi, and A.~Zisserman, ``Deep features for text
  spotting,'' in {\em Proc. Eur. Conf. Comp. Vis.}, 2014.

\bibitem{su2014accurate}
B.~Su and S.~Lu, ``Accurate scene text recognition based on recurrent neural
  network,'' in {\em Proc. Asian Conf. Comp. Vis.}, 2014.

\bibitem{su2017accurate}
B.~Su and S.~Lu, ``Accurate recognition of words in scenes without character
  segmentation using recurrent neural network,'' {\em Pattern Recogn.},
  vol.~63, pp.~397--405, 2017.

\bibitem{he2016reading}
P.~He, W.~Huang, Y.~Qiao, C.~C. Loy, and X.~Tang, ``Reading scene text in deep
  convolutional sequences,'' in {\em Proc. {AAAI} Conf. Artificial Intell.},
  2016.

\bibitem{yang2017learning}
X.~Yang, D.~He, Z.~Zhou, D.~Kifer, and C.~L. Giles, ``Learning to read
  irregular text with attention mechanisms.,'' in {\em Proc. Int. Joint Conf.
  Artificial Intell.}, 2017.

\bibitem{cheng2018aon}
Z.~Cheng, Y.~Xu, F.~Bai, Y.~Niu, S.~Pu, and S.~Zhou, ``Aon: Towards
  arbitrarily-oriented text recognition,'' in {\em Proc. IEEE Conf. Comp. Vis.
  Patt. Recogn.}, 2018.

\bibitem{zhan2019esir}
F.~Zhan and S.~Lu, ``{ESIR}: End-to-end scene text recognition via iterative
  image rectification,'' in {\em Proc. IEEE Conf. Comp. Vis. Patt. Recogn.},
  2019.

\bibitem{howard2017mobilenets}
A.~G. Howard, M.~Zhu, B.~Chen, D.~Kalenichenko, W.~Wang, T.~Weyand,
  M.~Andreetto, and H.~Adam, ``Mobilenets: Efficient convolutional neural
  networks for mobile vision applications,'' {\em arXiv preprint
  arXiv:1704.04861}, 2017.

\bibitem{lecun1998gradient}
Y.~LeCun, L.~Bottou, Y.~Bengio, and P.~Haffner, ``Gradient-based learning
  applied to document recognition,'' {\em Proceedings of the IEEE}, vol.~86,
  no.~11, pp.~2278--2324, 1998.

\bibitem{ioffe2015batch}
S.~Ioffe and C.~Szegedy, ``Batch normalization: Accelerating deep network
  training by reducing internal covariate shift,'' {\em arXiv preprint
  arXiv:1502.03167}, 2015.

\bibitem{lin2017feature}
T.-Y. Lin, P.~Doll{\'a}r, R.~Girshick, K.~He, B.~Hariharan, and S.~Belongie,
  ``Feature pyramid networks for object detection,'' in {\em Proc. IEEE Conf.
  Comp. Vis. Patt. Recogn.}, 2017.

\bibitem{vatti1992generic}
B.~R. Vatti, ``A generic solution to polygon clipping,'' {\em Communications of
  the ACM}, vol.~35, no.~7, pp.~56--63, 1992.

\bibitem{milletari2016v}
F.~Milletari, N.~Navab, and S.-A. Ahmadi, ``V-net: Fully convolutional neural
  networks for volumetric medical image segmentation,'' in {\em Int. Conf. 3D
  Vision}, 2016.

\bibitem{shrivastava2016training}
A.~Shrivastava, A.~Gupta, and R.~Girshick, ``Training region-based object
  detectors with online hard example mining,'' in {\em Proc. IEEE Conf. Comp.
  Vis. Patt. Recogn.}, 2016.

\bibitem{xu2019textfield}
Y.~Xu, Y.~Wang, W.~Zhou, Y.~Wang, Z.~Yang, and X.~Bai, ``Textfield: Learning a
  deep direction field for irregular scene text detection,'' {\em {IEEE} Trans.
  Image Process.}, vol.~28, no.~11, pp.~5566--5579, 2019.

\bibitem{baek2019character}
Y.~Baek, B.~Lee, D.~Han, S.~Yun, and H.~Lee, ``Character region awareness for
  text detection,'' in {\em Proc. IEEE Conf. Comp. Vis. Patt. Recogn.}, 2019.

\bibitem{zhang2019look}
C.~Zhang, B.~Liang, Z.~Huang, M.~En, J.~Han, E.~Ding, and X.~Ding, ``Look more
  than once: An accurate detector for text of arbitrary shapes,'' in {\em Proc.
  IEEE Conf. Comp. Vis. Patt. Recogn.}, 2019.

\bibitem{yao2012detecting}
C.~Yao, X.~Bai, W.~Liu, Y.~Ma, and Z.~Tu, ``Detecting texts of arbitrary
  orientations in natural images,'' in {\em Proc. IEEE Conf. Comp. Vis. Patt.
  Recogn.}, 2012.

\bibitem{lyu2018multi}
P.~Lyu, C.~Yao, W.~Wu, S.~Yan, and X.~Bai, ``Multi-oriented scene text
  detection via corner localization and region segmentation,'' {\em arXiv
  preprint arXiv:1802.08948}, 2018.

\bibitem{yao2014unified}
C.~Yao, X.~Bai, and W.~Liu, ``A unified framework for multioriented text
  detection and recognition,'' {\em {IEEE} Trans. Image Process.}, vol.~23,
  no.~11, pp.~4737--4749, 2014.

\bibitem{shi2017icdar2017}
B.~Shi, C.~Yao, M.~Liao, M.~Yang, P.~Xu, L.~Cui, S.~Belongie, S.~Lu, and
  X.~Bai, ``{ICDAR}2017 competition on reading {C}hinese text in the wild
  (rctw-17),'' in {\em Proc. Int. Conf. Document Analysis Recogn.}, 2017.

\bibitem{synthtext}
A.~Gupta, A.~Vedaldi, and A.~Zisserman, ``Synthetic data for text localisation
  in natural images,'' in {\em Proc. IEEE Conf. Comp. Vis. Patt. Recogn.},
  2016.

\bibitem{veit2016coco}
A.~Veit, T.~Matera, L.~Neumann, J.~Matas, and S.~Belongie, ``Coco-text: Dataset
  and benchmark for text detection and recognition in natural images,'' {\em
  arXiv preprint arXiv:1601.07140}, 2016.

\bibitem{nayef2017icdar2017}
N.~Nayef, F.~Yin, I.~Bizid, H.~Choi, Y.~Feng, D.~Karatzas, Z.~Luo, U.~Pal,
  C.~Rigaud, J.~Chazalon, {\em et~al.}, ``{ICDAR2017} robust reading challenge
  on multi-lingual scene text detection and script identification-rrc-mlt,'' in
  {\em Proc. Int. Conf. Document Analysis Recogn.}, 2017.

\bibitem{deng2009imagenet}
J.~Deng, W.~Dong, R.~Socher, L.-J. Li, K.~Li, and L.~Fei-Fei, ``{ImageNet}: A
  large-scale hierarchical image database,'' in {\em Proc. IEEE Conf. Comp.
  Vis. Patt. Recogn.}, 2009.

\bibitem{kingma2014adam}
D.~P. Kingma and J.~Ba, ``Adam: A method for stochastic optimization,'' {\em
  arXiv preprint arXiv:1412.6980}, 2014.

\bibitem{zhao2017pyramid}
H.~Zhao, J.~Shi, X.~Qi, X.~Wang, and J.~Jia, ``Pyramid scene parsing network,''
  in {\em Proc. IEEE Conf. Comp. Vis. Patt. Recogn.}, 2017.

\bibitem{rrpn}
J.~Ma, W.~Shao, H.~Ye, L.~Wang, H.~Wang, Y.~Zheng, and X.~Xue,
  ``Arbitrary-oriented scene text detection via rotation proposals,'' {\em
  {IEEE} Trans. Multimedia}, vol.~20, no.~11, pp.~3111--3122, 2018.

\bibitem{sun2018textnet}
Y.~Sun, C.~Zhang, Z.~Huang, J.~Liu, J.~Han, and E.~Ding, ``Textnet: Irregular
  text reading from images with an end-to-end trainable network,'' in {\em
  Proc. Asian Conf. Comp. Vis.}, 2018.

\bibitem{deepreg}
W.~He, X.-Y. Zhang, F.~Yin, and C.-L. Liu, ``Deep direct regression for
  multi-oriented scene text detection,'' in {\em Proc. IEEE Int. Conf. Comp.
  Vis.}, 2017.

\bibitem{he2017single}
P.~He, W.~Huang, T.~He, Q.~Zhu, Y.~Qiao, and X.~Li, ``Single shot text detector
  with regional attention,'' in {\em Proc. IEEE Int. Conf. Comp. Vis.}, 2017.

\bibitem{rrd}
M.~Liao, Z.~Zhu, B.~Shi, G.-s. Xia, and X.~Bai, ``Rotation-sensitive regression
  for oriented scene text detection,'' in {\em Proc. IEEE Conf. Comp. Vis.
  Patt. Recogn.}, 2018.

\bibitem{mcn}
Z.~Liu, G.~Lin, S.~Yang, J.~Feng, W.~Lin, and W.~L. Goh, ``Learning markov
  clustering networks for scene text detection,'' {\em Proc. IEEE Conf. Comp.
  Vis. Patt. Recogn.}, 2018.

\bibitem{liu2018path}
S.~Liu, L.~Qi, H.~Qin, J.~Shi, and J.~Jia, ``Path aggregation network for
  instance segmentation,'' in {\em Proc. IEEE Conf. Comp. Vis. Patt. Recogn.},
  2018.

\bibitem{ghiasi2019fpn}
G.~Ghiasi, T.-Y. Lin, and Q.~V. Le, ``{NAS-FPN}: Learning scalable feature
  pyramid architecture for object detection,'' in {\em Proc. IEEE Conf. Comp.
  Vis. Patt. Recogn.}, 2019.

\end{thebibliography}
	}

\end{document}